\documentclass{nature} % Nature class

\newif\ifdraft
\draftfalse

\usepackage[T1]{fontenc}
\usepackage[utf8]{inputenc}
\usepackage{textgreek}

\usepackage{helvet}    % Load Helvetica (Arial-like) font
\usepackage{sectsty}   % Allows font changes to section headers
\allsectionsfont{\sffamily}  % Apply Helvetica (sans-serif) to all section headers
\usepackage{titlesec}

\usepackage{newtxtext,newtxmath}    % new version of font
\usepackage{amsmath, amssymb, bm} % math
\usepackage{amsthm}               % theorem environments
\usepackage{graphicx}             % figures
\usepackage{xspace}               % smart spaces in macros
\usepackage{booktabs}             % tables
\usepackage{multirow}
\usepackage{pdflscape}            % landscape pages (optional)
\usepackage[left=1in,right=1in,bottom=1.1in,top=1in]{geometry}
\usepackage{enumitem}             % compact lists
\usepackage[table]{xcolor}        % color (single load)
\usepackage[dvipsnames]{xcolor}   % additional colors
\usepackage{siunitx}              % SI units for methods
\usepackage{textcomp}
\usepackage[normalem]{ulem}       % \emph unaffected; allows \sout for drafts
\usepackage[ruled,vlined,linesnumbered]{algorithm2e} % algorithms
\usepackage[innercaption]{sidecap} % side captions
\usepackage{lineno}               % line numbers (optional)
\usepackage{verbatim}             % verbatim (kept)
\let\spacing\relax
\usepackage{setspace}             % spacing (we’ll use environments)
\usepackage{xr}                   % external references
\usepackage[colorlinks,citecolor=blue,urlcolor=blue]{hyperref} % load near last
\usepackage{xr-hyper}
\usepackage{float}
\usepackage{soul}

\usepackage[table]{xcolor} % colour support for tables
\usepackage{tabularx}      % for automatic column widths
\usepackage{authaffil}         % authors and affiliations

\usepackage[natbib=true, citestyle=numeric-comp,
            bibstyle=nature, sorting=none,
            maxbibnames=6, doi=true, url=false, isbn=false, date=year]{biblatex}
\AtEveryBibitem{%
  \clearfield{note}%
  \clearlist{note}%
  \clearfield{language}%
  \clearlist{language}%
}

\usepackage[font=footnotesize, labelfont={sf,bf}]{caption}
\DeclareCaptionLabelFormat{extended_data}{\textbf{Extended Data #1 #2: }}
\DeclareCaptionLabelFormat{supplementary}{\textsf{\textbf{Supplementary #1 #2: }}} % applied to SI floats only

\graphicspath{{FIG/}}

\newcommand{\xhdr}[1]{\vspace{1.2mm}\noindent\textsf{\textbf{{#1.}}}\xspace}

\titlespacing*{\section}{0pt}{1em}{0em}
\titlespacing*{\paragraph}{0pt}{0pt}{1em}

\newcommand{\eg}{\emph{e.g.}\xspace}
\newcommand{\ie}{\emph{i.e.}\xspace}

\newcommand{\textsbf}[1]{\textsf{\textbf{#1}}}

\newcommand\name{\textsc{MedLog}\xspace}
\newcommand\beacon{\textsc{BEACON}\xspace}

\definecolor{mlHeader}{HTML}{000000}
\definecolor{mlModel}{HTML}{000000}
\definecolor{mlUser}{HTML}{000000}
\definecolor{mlTarget}{HTML}{000000}
\definecolor{mlInputs}{HTML}{000000}
\definecolor{mlArtifacts}{HTML}{000000}
\definecolor{mlOutputs}{HTML}{000000}
\definecolor{mlOutcomes}{HTML}{000000}
\definecolor{mlFeedback}{HTML}{000000}

\DeclareRobustCommand{\header}[1]{\textbf{\textcolor{mlHeader}{#1}}}
\DeclareRobustCommand{\model}[1]{\textbf{\textcolor{mlModel}{#1}}}
\DeclareRobustCommand{\user}[1]{\textbf{\textcolor{mlUser}{#1}}}
\DeclareRobustCommand{\target}[1]{\textbf{\textcolor{mlTarget}{#1}}}
\DeclareRobustCommand{\inputs}[1]{\textbf{\textcolor{mlInputs}{#1}}}
\DeclareRobustCommand{\artifacts}[1]{\textbf{\textcolor{mlArtifacts}{#1}}}
\DeclareRobustCommand{\outputs}[1]{\textbf{\textcolor{mlOutputs}{#1}}}
\DeclareRobustCommand{\outcomes}[1]{\textbf{\textcolor{mlOutcomes}{#1}}}
\DeclareRobustCommand{\feedback}[1]{\textbf{\textcolor{mlFeedback}{#1}}}

\newcommand{\mlitemheader}{\item[\textsf{\textbf{\textcolor{mlHeader}{1.}}}]}
\newcommand{\mlitemmodel}{\item[\textsf{\textbf{\textcolor{mlModel}{2.}}}]}
\newcommand{\mlitemuser}{\item[\textsf{\textbf{\textcolor{mlUser}{3.}}}]}
\newcommand{\mlitemtarget}{\item[\textsf{\textbf{\textcolor{mlTarget}{4.}}}]}
\newcommand{\mliteminputs}{\item[\textsf{\textbf{\textcolor{mlInputs}{5.}}}]}
\newcommand{\mlitemartifacts}{\item[\textsf{\textbf{\textcolor{mlArtifacts}{6.}}}]}
\newcommand{\mlitemoutputs}{\item[\textsf{\textbf{\textcolor{mlOutputs}{7.}}}]}
\newcommand{\mlitemoutcomes}{\item[\textsf{\textbf{\textcolor{mlOutcomes}{8.}}}]}
\newcommand{\mlitemfeedback}{\item[\textsf{\textbf{\textcolor{mlFeedback}{9.}}}]}

\title{
\begin{center}
A global log for medical AI
\vspace{-12mm}
\end{center}
}

\DefineAffiliation{HMS}{Department of Biomedical Informatics, Harvard Medical School, Boston, MA, USA}
\DefineAffiliation{OXENG}{Department of Engineering Science, University of Oxford, Oxford, UK}
\DefineAffiliation{OXETHICS}{Institute for Ethics in AI, University of Oxford, Oxford, UK}
\DefineAffiliation{COSMOS}{Cosmos Institute, Austin, TX, USA}
\DefineAffiliation{BERK}{The Ivan and Francesca Berkowitz Family Living Laboratory Collaboration at \\ Harvard Medical School and Clalit Research Institute, Boston, MA, USA}
\DefineAffiliation{UCSDBMI}{Division of Biomedical Informatics, Department of Medicine,\\ University of California, San Diego, La Jolla, CA, USA}
\DefineAffiliation{JACOBS}{Joan and Irwin Jacobs Center for Health Innovation, UC San Diego Health, San Diego, CA, USA}
\DefineAffiliation{OUCRU}{Oxford University Clinical Research Unit, Ho Chi Minh City, Vietnam}
\DefineAffiliation{MSWIN}{The Windreich Department of Artificial Intelligence and Human Health,\\ Icahn School of Medicine at Mount Sinai, New York, NY, USA}
\DefineAffiliation{MSHPI}{The Hasso Plattner Institute for Digital Health at Mount Sinai, Icahn School of Medicine\\ at Mount Sinai and Mount Sinai Health System, New York City, NY, USA}
\DefineAffiliation{ETHCS}{Department of Computer Science, ETH Zurich, Z\"urich, Switzerland}
\DefineAffiliation{BERN}{Department of Intensive Care Medicine, University Hospital, University of Bern, Bern, Switzerland}
\DefineAffiliation{BIDMC}{Division of General Internal Medicine, Department of Medicine,\\ Beth Israel Deaconess Medical Center, Boston, MA, USA}
\DefineAffiliation{DEEPMIND}{Google DeepMind, London, UK}
\DefineAffiliation{ETHHST}{Department of Health Sciences and Technology, ETH Zurich, Z\"urich, Switzerland}
\DefineAffiliation{MSAIAL}{The Mount Sinai AI Assurance Lab, Mount Sinai Health System, New York, NY}
\DefineAffiliation{BHAM}{University of Birmingham, Birmingham, UK}
\DefineAffiliation{PATH}{PATH, Seattle, WA, USA}
\DefineAffiliation{SEACHI}{Office of the CEO, Seattle Children's Hospital, Seattle, WA, USA}
\DefineAffiliation{UCSDMED}{Department of Medicine, University of California, San Diego, La Jolla, CA, USA}
\DefineAffiliation{UCSDPED}{Department of Pediatrics, University of California, San Diego, La Jolla, CA, USA}
\DefineAffiliation{KAISER}{Kaiser Foundation Health and Hospitals, Oakland, CA, USA}
\DefineAffiliation{EPATIENT}{e-Patient Dave, LLC, Nashua, NH, USA}
\DefineAffiliation{WHOEURO}{Regional Office for Europe, World Health Organization, Copenhagen, Denmark}
\DefineAffiliation{MALTA}{University of Malta, Msida, Malta}
\DefineAffiliation{OXTROP}{Centre for Tropical Medicine and Global Health, University of Oxford, Oxford, UK}
\DefineAffiliation{HIMSS}{Healthcare Information and Management Systems Society, Chicago, IL, USA}
\DefineAffiliation{CLALIT}{Clalit Research Institute, Innovation Division, Clalit Health Services, Ramat Gan, Israel}
\DefineAffiliation{MSR}{Microsoft Research, Redmond, WA, USA}
\DefineAffiliation{GATES}{Gates Foundation, Seattle, WA, USA}
\DefineAffiliation{CHIP}{Computational Health Informatics Program, Boston Children's Hospital, Boston, MA, USA}
\DefineAffiliation{HMSPED}{Department of Pediatrics, Harvard Medical School, Boston, MA, USA}
\DefineAffiliation{RADY}{Rady School of Management, University of California, San Diego, CA, USA}
\DefineAffiliation{LUMS}{Department of Computer Science, Lahore University of Management Sciences, Lahore, Pakistan}
\DefineAffiliation{AWAAZ}{Awaaz-e-Sehat, Lahore, Pakistan}
\DefineAffiliation{STANTDS}{Technology and Digital Solutions, Stanford Healthcare, Palo Alto, CA, USA}
\DefineAffiliation{STANMED}{Department of Medicine, Stanford University, Stanford, CA, USA}
\DefineAffiliation{STANCERC}{Clinical Excellence Research Center, Stanford University, Stanford, CA, USA}
\DefineAffiliation{BGUCS}{Faculty of Computer and Information Science, Ben Gurion University of the Negev, Be'er Sheva, Israel}
\DefineAffiliation{BGUPH}{Faculty of Health Sciences, School of Public Health, Ben Gurion University of the Negev, Be'er Sheva, Israel}
\DefineAffiliation{PENDA}{Penda Health, Nairobi, Kenya}
\DefineAffiliation{EPIC}{Epic Systems Corporation, Verona, WI, USA}
\DefineAffiliation{TSINGHUA}{Tsinghua Medicine, Tsinghua University, Beijing, China}
\DefineAffiliation{SERI}{Singapore Eye Research Institute, Singapore National Eye Centre, Singapore}
\DefineAffiliation{BERI}{Beijing Visual Science and Translational Eye Research Institute (BERI), School of Clinical Medicine, Beijing Tsinghua Changgung Hospital, Beijing, China}
\DefineAffiliation{MSCBI}{The Charles Bronfman Institute for Personalized Medicine, Icahn School of Medicine at Mount Sinai\\ and Mount Sinai Health System, New York City, NY, USA}
\DefineAffiliation{MSICCM}{Institute for Critical Care Medicine, Icahn School of Medicine at Mount Sinai, New York, NY}
\DefineAffiliation{MSMINDICH}{Mindich Child Health and Development Institute and the Departments of Pediatrics and Genetics \& Genomic Sciences, Icahn School of Medicine at Mount Sinai, New York, NY, USA}
\DefineAffiliation{OXGHR}{Centre for Global Health Research, Nuffield Department of Medicine, University of Oxford, Oxford, UK}
\DefineAffiliation{OXSUZHOU}{Oxford Suzhou Centre for Advanced Research, University of Oxford, Suzhou, Jiangsu, China}
\DefineAffiliation{KEMPNER}{Kempner Institute for the Study of Natural and Artificial Intelligence at Harvard University, Allston, MA, USA}
\DefineAffiliation{BROAD}{Broad Institute of MIT and Harvard, Cambridge, MA, USA}
\DefineAffiliation{HDSI}{Harvard Data Science Initiative, Cambridge, MA, USA}

\AddAuthor[][0000-0003-1420-1236]{Ayush~Noori}{HMS,OXENG,OXETHICS,COSMOS,BERK}
\AddAuthor[*]{Aaron~E.~Boussina}{UCSDBMI,JACOBS}
\AddAuthor[*]{Hai~Ho~Bich}{OUCRU}
\AddAuthor[*]{James~Anibal}{OXENG}
\AddAuthor[*]{Julia~Maslinski}{MSWIN,MSHPI}
\AddAuthor[*]{Manuel~Burger}{ETHCS}
\AddAuthor[*]{Martin~Faltys}{BERN}
\AddAuthor{Adam~Rodman}{BIDMC}
\AddAuthor{Alan~Karthikesalingam}{DEEPMIND}
\AddAuthor{Alessandro~Blasimme}{ETHHST}
\AddAuthor{Annelia~Itwaru}{MSAIAL}
\AddAuthor{Ben~Kaplan}{MSAIAL}
\AddAuthor{Bilal~A.~Mateen}{BHAM,PATH}
\AddAuthor{Christopher~A.~Longhurst}{SEACHI,UCSDMED,UCSDPED}
\AddAuthor{Daniel~Yang}{KAISER}
\AddAuthor{Dave~deBronkart}{EPATIENT}
\AddAuthor{Effy~Vayena}{ETHHST}
\AddAuthor{Fedor~Sergeev}{ETHCS}
\AddAuthor{Gauden~Galea}{WHOEURO,MALTA}
\AddAuthor{Ha~Thi~Hai~Duong}{OUCRU,OXTROP}
\AddAuthor{Harold~F.~Wolf~III}{HIMSS}
\AddAuthor{Jacob~Waxman}{CLALIT}
\AddAuthor{Joerg~C.~Schefold}{BERN}
\AddAuthor{Joshua~C.~Mandel}{HMS,MSR}
\AddAuthor{Juliana~Rotich}{GATES}
\AddAuthor{Kenneth~D.~Mandl}{HMS,CHIP,HMSPED}
\AddAuthor{Lily~Poursoltan}{JACOBS,RADY}
\AddAuthor{Maryam~Mustafa}{LUMS,AWAAZ}
\AddAuthor{Melissa~Miles}{GATES}
\AddAuthor{Nigam~H.~Shah}{STANTDS,STANMED,STANCERC}
\AddAuthor{Noa~Dagan}{BERK,CLALIT,BGUCS}
\AddAuthor{Pavan~Bodanki}{MSAIAL}
\AddAuthor{Peter~Lee}{MSR}
\AddAuthor{Philipp~Koralus}{OXETHICS,COSMOS}
\AddAuthor{Prathamesh~Parchure}{MSAIAL}
\AddAuthor{Prem~Timsina}{MSAIAL,MSWIN}
\AddAuthor{Ran~D.~Balicer}{BERK,CLALIT,BGUPH}
\AddAuthor{Robert~Korom}{PENDA}
\AddAuthor{Scott~Mahoney}{GATES}
\AddAuthor{Seth~Hain}{EPIC}
\AddAuthor{Tien~Yin~Wong}{TSINGHUA,SERI,BERI}
\AddAuthor{Trevor~Mundel}{GATES}
\AddAuthor{Vivek~Natarajan}{DEEPMIND}
\AddAuthor[\dagger]{Ankit~Sakhuja}{MSAIAL,MSWIN,MSHPI,MSCBI,MSICCM}
\AddAuthor[\dagger]{Benjamin~Glicksberg}{MSWIN,MSHPI,MSMINDICH}
\AddAuthor[\dagger]{C.~Louise~Thwaites}{OUCRU,OXTROP,OXGHR}
\AddAuthor[\dagger]{Gunnar~R\"{a}tsch}{MSWIN}
\AddAuthor[\dagger]{Karandeep~Singh}{UCSDBMI,JACOBS}
\AddAuthor[\dagger]{David~A.~Clifton}{OXENG,OXSUZHOU}
\AddAuthor[\dagger,\ddagger][0000-0003-2192-5160]{Isaac~S.~Kohane}{HMS,BERK,CHIP}
\AddAuthor[\dagger,\ddagger][0000-0001-8530-7228]{Marinka~Zitnik}{HMS,BERK,KEMPNER,BROAD,HDSI}

\AddAffilNote{*}{These authors contributed equally as co-second authors and are ordered alphabetically.}
\AddAffilNote{\dagger}{These authors jointly supervised this work.}
\AddAffilNote{\ddagger}{Correspondence: \href{mailto:isaac_kohane@hms.harvard.edu}{isaac\_kohane@hms.harvard.edu}, \href{mailto:marinka@hms.harvard.edu}{marinka@hms.harvard.edu}.}

\author{{\small\begin{center}
    \PrintAuthors \\[2mm]
    {\footnotesize
      \name website: \url{https://medlogprotocol.ai}\\
      \name code: \url{https://github.com/mims-harvard/MedLog}%
    }\\[2mm]
    \PrintAffiliations \\[1mm]
    \PrintAffilNotes\\
    \vspace{-1em}
\end{center}}}

\begin{document}

\makeatletter
\let\savedauthorblock\@author
\makeatother
\maketitle

% ================
% Optional: line numbers during review
% ================
% \linenumbers
\renewcommand{\linelabel}[1]{}

% =========================
% MAIN TEXT
% =========================

\begin{refsection}

% =========================
% Abstract
% =========================
% \clearpage
\section*{Abstract}
\begin{spacing}{1}
\small
\begin{abstract}
\noindent Modern computer systems rely on syslog, a universal protocol that records critical events across heterogeneous infrastructure. Medicine's rapidly growing AI stack has no equivalent. As medicine deploys AI tools at scale, there is no standard way to record how, when, by whom, and for whom these models are used. Without such records, it is difficult to measure real-world performance and outcomes, detect adverse events, or identify bias and dataset drift. Here we introduce \name, a protocol for event-level logging of medical AI. Each time an AI model interacts with a human, another algorithm, or an automated workflow, \name creates a record. Each record contains nine core fields: header, model, user, target, inputs, artifacts, outputs, outcomes, and feedback.
We apply \name across four deployments in the US, Switzerland, and Vietnam: ICU deterioration prediction, tetanus progression monitoring from wearable signals, automated sepsis quality reporting, and patient attendance prediction. \name records capture model behavior, workflow interactions, and downstream outcomes, including AI performance degradation during severe weather events in patient attendance prediction and increased laboratory testing after ICU deterioration alerts.
\name limits the data footprint through risk-based sampling, lifecycle-aware retention policies, and write-behind caching, enabling deployment in low-resource settings. It also supports detailed traces for complex, agentic, or multi-stage workflows, creating a foundation for continuous monitoring, auditing, and improvement of medical AI.
\end{abstract}
\end{spacing}
\clearpage

% =========================
% Main text
% =========================
\begin{spacing}{1.3}

\section*{Main}
Artificial intelligence (AI) is rapidly entering clinical settings in a fragmented and largely unregulated manner. As of May 2025, at least 377 healthcare systems and providers in the U.S. have piloted or adopted 70 generative AI tools developed by 49 companies for clinical decision support, patient communication, documentation, claims processing, and healthcare administration \autocite{palmer_guide_2023, umeton_gpt-4_2024, landi_epic_2025}. The majority of American physicians report using AI technologies in clinical care \autocite{american_medical_association_physician_2025}. Similar trends are observed globally: 48\% of clinicians across 109 countries report AI use in their work \autocite{elsevier_clinician_2025}, and large-scale deployments are emerging across health systems in China \autocite{zeng_deepseeks_2025}.

This adoption reflects AI performance on benchmark evaluations, though most benchmarks use synthetic or controlled data. Large language models (LLMs) match or exceed physician performance in diagnostic reasoning \autocite{kanjee_accuracy_2023, goh_large_2024, cabral_clinical_2024}, clinical text summarization \autocite{van_veen_adapted_2024}, medical question answering \autocite{xie_preliminary_2024, katz_gpt_2024, singhal_large_2023, tu_towards_2024}, patient communication \autocite{ayers_comparing_2023, tu_towards_2025}, and multi-step reasoning tasks \autocite{brodeur_superhuman_2024}. Yet concerns about hallucinations, bias, and reliability remain \autocite{ahmad_creating_2023, kim_medical_2025, mittermaier_bias_2023, cross_bias_2024}, and real-world clinical performance is poorly understood \autocite{wornow_shaky_2023}. More realistic benchmarks are emerging \autocite{arora_healthbench_2025, bedi_medhelm_2025}, but systematic evaluation in deployed settings remains rare \autocite{shah_making_2019, youssef_external_2023, agweyu_retrospective_2025}.

Medical AI lacks consistent logging of model use. Without records of how models are used in practice, health systems cannot assess performance, detect failure modes, or measure clinical impact~\autocite{keyes_why_2026}. \linelabel{line:intro-ehr-audit-log-start} Audit logs of electronic health records (EHRs) provide timestamped records of EHR activity and have been used to study clinical workflows~\autocite{adler-milstein_ehr_2020, rule_using_2023, kannampallil_using_2023}, but do not capture AI model usage. \linelabel{line:intro-ehr-audit-log-end} Reporting frameworks, including TRIPOD+AI, STARD-AI, DECIDE-AI, SPIRIT-AI, and CONSORT-AI \autocite{collins_reporting_2019, collins_tripodai_2024, sounderajah_developing_2020, vasey_decide-ai_2021, cruz_rivera_guidelines_2020}, focus on model development and evaluation in controlled settings, and do not address continuous, event-level monitoring of deployed systems, including generative AI and agentic workflows. Several governance and evaluation frameworks have also been proposed \autocite{dagan_evaluation_2024, callahan_standing_2024, lekadir_future-ai_2025, ong_international_2025, epic_open_source_epic-open-sourceseismometer_2025, coalition_for_health_ai_assurance_2024}, along with targeted monitoring systems for specific deployments~\autocite{bedoya_framework_2022, umeton_gpt-4_2024, shah_adoption_2026}. No broadly adopted standard records each instance of AI use in clinical care~\autocite{feng_not_2025}.

This gap limits both the evaluation of medical AI in practice and the governance of its use. The FDA has emphasized that clinicians cannot feasibly oversee all outputs of generative AI systems \cite{Warraich2025-mu}, making systematic monitoring a requirement. In other domains, centralized logging protocols such as \texttt{syslog} record each system event across many connected services \autocite{lonvick_bsd_2001, gerhards_syslog_2009} (Supplementary Table~\ref{si:tab:syslog_medlog}), supporting real-time monitoring, root cause analysis, and auditing at scale \autocite{zhang_system_2023, forrester_research_inc_total_2024}. Medical AI has no equivalent standard.

\begin{figure}[t]
  \centering
  \includegraphics[width=0.99\textwidth]{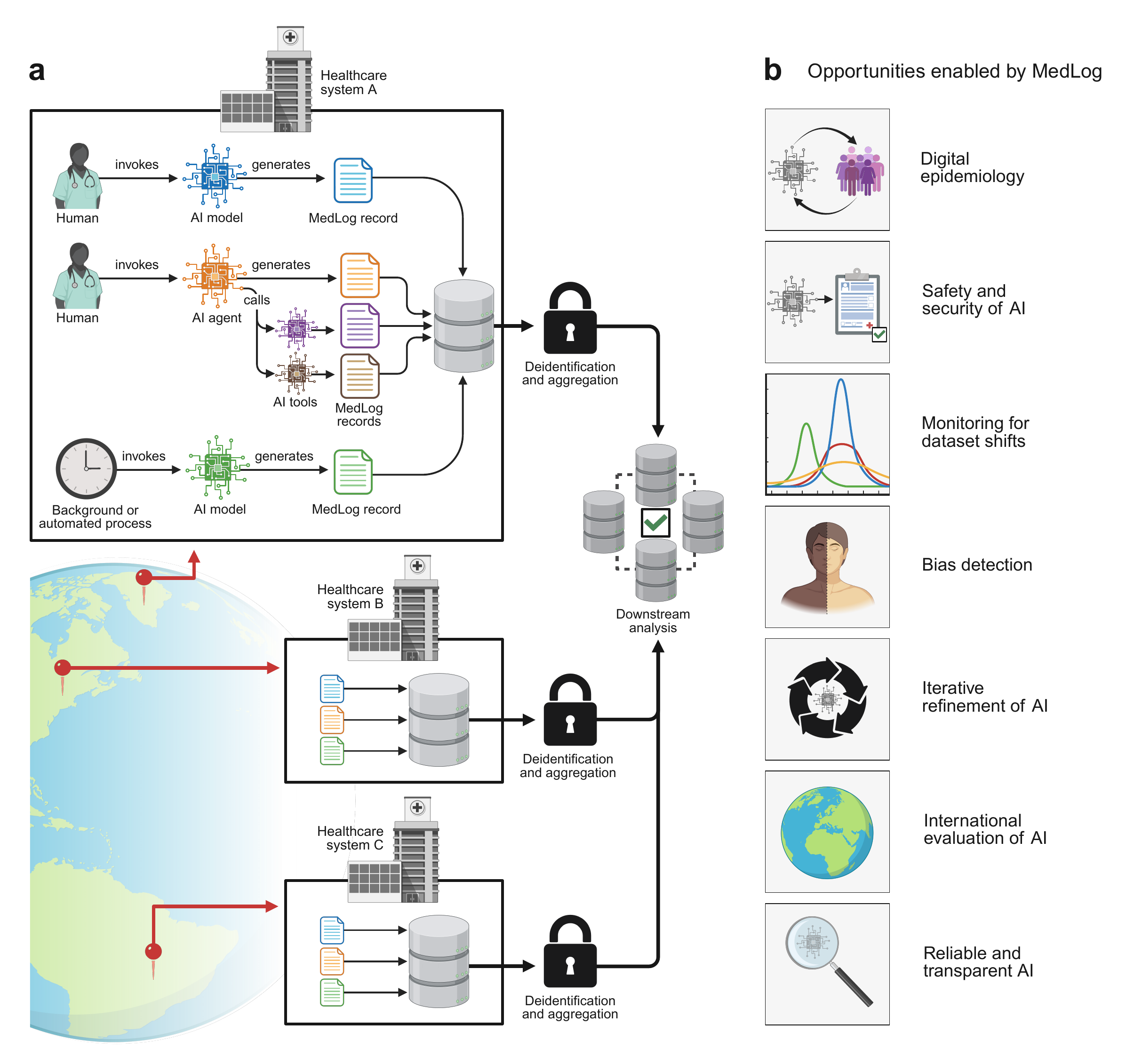}
  \caption{\textsbf{Overview of \name.} \textsbf{(a)} Timeline demonstrating that \name records are progressively built from a stream of messages. \textsbf{(b)} Examples of medical AI interactions that can be logged under the \name protocol, as well as the \name records they create. Full-length example records are available in Supplementary Figure~\ref{si:fig:examples}. \textsbf{(c)} We report four clinical deployments of \name across Switzerland, Vietnam, and the United States.}
  \label{fig:elements}
\end{figure}

Here we introduce \name, a protocol for event-level logging of medical AI. \name specifies a schema for each model invocation, including inputs, outputs, intermediate artifacts, and, when available, clinical outcomes and user feedback. It records both single-step interactions and multi-stage agentic workflows by linking events over time (Figure~\ref{fig:elements}a). \linelabel{line:medlog-scope-r3-start} \name covers any AI process that uses health data and can affect patient outcomes: interactions between models and patients, clinicians, administrators, and other stakeholders; background services such as batch inference, autonomous triage, claim routing, and continuous monitoring; and AI-AI exchanges within agentic workflows and orchestration frameworks (Figure~\ref{fig:elements}b).\linelabel{line:medlog-scope-r3-end}

We evaluate \name across four clinical deployments spanning intensive care monitoring, infectious disease severity prediction, hospital quality reporting, and patient attendance prediction (Figure~\ref{fig:elements}c, Table~\ref{tab:pilots}). Across these deployments, \name makes model behavior visible in practice, revealing patterns that offline evaluation does not capture, including temporal failure modes, workflow-dependent variability, interactions between model outputs and clinician behavior, and performance degradation during severe weather events. By standardizing how health systems record AI use and link it to outcomes, \name provides the infrastructure needed to measure real-world performance, detect failures, and guide deployment at scale.

% FIGURE 1

\section*{A \name record captures each model invocation}\label{sec:proposal}
\name complements model cards and data sheets, which document models and datasets before deployment. Model cards describe properties of an AI model, including its architecture, training objective, evaluation results, intended use, and known limitations \autocite{mitchell_model_2019, openai_gpt-4o_2024, meta_llama_2024}. Data sheets describe how datasets were collected, processed, and characterized, including demographic composition and known limitations \autocite{gebru_datasheets_2021}. These documents do not record how a model is used after deployment. \name addresses this gap by recording each model invocation, the information available at that invocation, the output produced, and, when feasible,  clinical or operational outcomes.

\linelabel{line:future-proofing-start} \name captures AI use at the level of an inference call, covering simple ``prompt to model to response'' systems as well as multi-stage workflows that use retrieval, tools, rubric-based evaluation, or agentic orchestration. \linelabel{line:multiturn-1-start} Events can be linked within the same run or episode, allowing \name to represent both single interactions and extended workflows (Figure~\ref{fig:elements}a). \linelabel{line:multiturn-1-end} Each \name record corresponds to one model invocation and contains the following fields.\linelabel{line:future-proofing-end} 

\begin{enumerate}

    \mlitemheader \textsf{\header{Header.}} The \name record header consists of provenance information, execution context, and system metadata available at inference time, including server identifiers, timestamps of model invocation and input retrieval, and a stable event identifier for the newly-created \name record. \linelabel{line:multiturn-2-start} Akin to \texttt{PROCID} in \texttt{syslog}, this field \linelabel{line:rfc-language-1} may also optionally include identifiers for run- or episode-level linkage (\eg, \texttt{run\_id}, \texttt{parent\_event\_id}) \autocite{gerhards_syslog_2009}. These identifiers allow grouping events across multi-stage or agentic workflows without imposing a specific orchestration architecture. \linelabel{line:multiturn-2-end} To ensure interoperability, the header \linelabel{line:rfc-language-2} must include the version of the \name protocol specification that the record complies with.
    
    \mlitemmodel \textsf{\model{Model instance.}} Stable identifiers of the AI model and version, with references to its model card and data sheet. When no data sheet is available, this field \linelabel{line:rfc-language-3}  records the training data version and any databases queried for retrieval-augmented generation or other knowledge injection. Test-time edits to the model \linelabel{line:rfc-language-4} are also recorded.
    
    \mlitemuser \textsf{\user{User identity.}} The technical process, service, or workflow that invokes the model call. At a minimum, the identifier of the immediate calling process \linelabel{line:rfc-language-5} must be logged. When possible, the upstream users who initiated the call \linelabel{line:rfc-language-6} should also be recorded as a provenance chain. Human users, such as clinicians or patients, should be identified by their EHR identifiers, including National Provider Identifier (NPI) or medical record number (MRN). Users may also be algorithms or automated systems — AI agents or scheduled jobs that trigger models for tasks such as risk calculation or triage \autocite{williams_use_2024, hinson_multisite_2022, gao_empowering_2024}. The level of user detail will vary across settings; attribution to an individual clinician is more straightforward in outpatient care than in inpatient team-based workflows.
    
    \mlitemtarget \textsf{\target{Target identity.}} A reference to the entity about which the model produces output, when applicable, for example, a patient ID for clinical predictions or a claim ID for administrative tasks. This field is optional for models that do not produce outputs about discrete targets.
    
    \mliteminputs \textsf{\inputs{Inputs.}} The input data provided to the model. For structured predictive models, this includes feature vectors or structured fields. For generative models such as LLMs, it should include prompts, instructions, and any relevant environmental variables. When inputs are too large to log directly, such as imaging or genomic data, stable identifiers sufficient to retrieve the input data retrospectively \linelabel{line:rfc-language-7} should be recorded.

    \mlitemartifacts \textsf{\artifacts{Internal artifacts.}}  Artifacts generated during inference for technical researchers and MLOps engineers. This field \linelabel{line:rfc-language-8} may include reasoning traces, such as chain-of-thought \autocite{wei_chain--thought_2022}, tree-of-thought \autocite{yao_tree_2023}, or graph-of-thought prompting paths \autocite{besta_graph_2024}; context retrieved during retrieval-augmented generation \autocite{ng_rag_2024}; agent interaction traces, such as iterative hypothesis testing or round-table discussions among AI agents \autocite{gao_empowering_2024, chen_reconcile_2024}; uncertainty estimates, such as confidence scores, prediction intervals, generation quality metrics, or entropy-based measures; and interpretability artifacts, including attribution maps, feature-importance scores, or saliency maps \autocite{degrave_ai_2021, saraswat_explainable_2022}. For models that update continuously or episodically, including self-evolving models or agents \autocite{gao_survey_2025} with Bayesian updating, persistent memory \autocite{li_memos_2025}, lifelong learning \autocite{zheng_lifelong_2025}, or dynamic routing or reconfiguration \autocite{guo_dynamic_2025}, this field may also record relevant model states or memory snapshots. These records allow retrospective reconstruction of the model configuration at the time of use with greater detail than using only the version identifier from the Model instance field.

    \mlitemoutputs \textsf{\outputs{Patient- or clinician-facing outputs.}} The outputs intended for human users: predictive outputs such as labels, risk scores, or forecasts, with associated confidence measures; generative outputs such as text, images, or videos; explanations, reasoning traces or rationales distilled from internal artifacts; and recommendations generated by single or multi-agent systems. Any triage levels or risk scores that determine whether a \name record is flagged for human review \linelabel{line:rfc-language-9} should also be recorded in this field.
    
    \mlitemoutcomes \textsf{\outcomes{Outcomes.}} When feasible, records of clinical actions or patient outcomes linked to the model recommendation, for example, whether a suggested therapy was administered and the observed clinical result. Outcome linkage is often indirect, the connection between recommendation and action may be delayed, and relevant data may reside outside the immediate AI workflow. Even when outcome data are incomplete, partial linkage remains valuable for post-deployment surveillance, epidemiological analysis, and model improvement. Outcomes \linelabel{line:rfc-language-10} may also include traces of how patients or clinicians interact with the EHR after viewing AI outputs, as recorded in EHR audit logs \autocite{kannampallil_using_2022}. Retrospective outcomes may be linked to a \name record using provider attestations, temporal proximity, trial emulations, or automated queries, and the strength of each linkage may be recorded to support tiered evidence standards for outcome attribution.
    
    \mlitemfeedback \textsf{\feedback{User feedback.}} Any feedback provided by users, whether structured ratings or free-text comments, \linelabel{line:rfc-language-11} should be recorded to support model refinement and user experience improvements.
\end{enumerate}

\name supports two levels of logging: compact records for system-wide monitoring and detailed traces for workflows that require reconstruction or review. Core fields are designed for broad adoption across medical AI deployments. The Internal artifacts field is optional and can record execution traces under predefined capture policies, such as continuous logging, random sampling, risk-triggered collection, or enhanced tracing after major updates and during phased deployment.

% TABLE 1
% ===== Pilot summary macros =====

% Bern, Switzerland
\newcommand{\BernDurationDays}{114}
\newcommand{\BernPatients}{212}
\newcommand{\BernMale}{148}
\newcommand{\BernFemale}{64}
\newcommand{\BernMedianAge}{68.00}
\newcommand{\BernAgeIQRLow}{55.80}
\newcommand{\BernAgeIQRHigh}{77.00}
\newcommand{\BernRecords}{223,840}

% Ho Chi Minh City, Vietnam
\newcommand{\VietnamDurationDays}{289}
\newcommand{\VietnamPatients}{15}
\newcommand{\VietnamMale}{14}
\newcommand{\VietnamFemale}{1}
\newcommand{\VietnamMedianAge}{49.94}
\newcommand{\VietnamAgeIQRLow}{47.03}
\newcommand{\VietnamAgeIQRHigh}{60.54}
\newcommand{\VietnamRecords}{3,406}

% San Diego, California, USA
\newcommand{\SanDiegoDurationDays}{89}
\newcommand{\SanDiegoPatients}{60}
\newcommand{\SanDiegoMale}{28}
\newcommand{\SanDiegoFemale}{32}
\newcommand{\SanDiegoMedianAge}{64.00}
\newcommand{\SanDiegoAgeIQRLow}{54.00}
\newcommand{\SanDiegoAgeIQRHigh}{71.25}
\newcommand{\SanDiegoRecords}{3,766}

% New York, New York, USA
\newcommand{\NewYorkDurationDays}{244}
\newcommand{\NewYorkPatients}{791,319}
\newcommand{\NewYorkMale}{324,294}
\newcommand{\NewYorkFemale}{466,626}
\newcommand{\NewYorkOther}{399}
\newcommand{\NewYorkMedianAge}{54.00}
\newcommand{\NewYorkAgeIQRLow}{36.00}
\newcommand{\NewYorkAgeIQRHigh}{68.00}
\newcommand{\NewYorkRecords}{2,914,264}
\newcommand{\NewYorkEncounters}{2,914,264}

% Brand colours (tweak to taste)
\definecolor{HeaderBG}{HTML}{003366}   % deep navy
\definecolor{HeaderFG}{HTML}{FFFFFF}   % white
\definecolor{AttrBG}  {HTML}{E3E8F0}   % very light slate
\definecolor{RowAlt}  {HTML}{F7F9FC}   % ultra-light gray
\arrayrulecolor{gray!55}               % light gray rules

% Shorthand for left-column cells
\newcommand{\attrcell}[1]{\cellcolor{AttrBG}\textbf{#1}}
\newcommand{\tightrule}{\specialrule{\lightrulewidth}{0pt}{0pt}}
\setlength{\tabcolsep}{6pt} 

% Define a centered X column type
\newcommand{\head}[1]{\multicolumn{1}{c}{\textbf{#1}}}

% TABLE 1
\begin{table}[t]
  \centering
  \rowcolors{3}{RowAlt}{white}
  \begin{tabularx}{\linewidth}
  {@{\hspace*{6pt}}
    >{\raggedright\arraybackslash}p{3.5cm}
    X X X X
   @{\hspace*{6pt}}}
    % --- header ---
    \rowcolor{HeaderBG}
    \textcolor{HeaderFG}{\textbf{Location}} &
    \textcolor{HeaderFG}{\textbf{Duration}} &
    \textcolor{HeaderFG}{\textbf{Number of\newline patients}} &
    \textcolor{HeaderFG}{\textbf{Median age\newline (IQR)}} &
    \textcolor{HeaderFG}{\textbf{No. \name\newline records}} \\[0.2em]
    \tightrule
    
    % --- rows ---
    \attrcell{Bern,\newline Switzerland} &
      \BernDurationDays\ days &
      \BernPatients\newline(\BernMale\ M, \BernFemale\ F) &
      \BernMedianAge\newline(\BernAgeIQRLow\ -- \BernAgeIQRHigh) &
      \BernRecords \\ \tightrule

    \attrcell{Ho Chi Minh City, Vietnam} &
      \VietnamDurationDays\ days &
      \VietnamPatients\newline(\VietnamMale\ M, \VietnamFemale\ F) &
      \VietnamMedianAge\newline(\VietnamAgeIQRLow\ -- \VietnamAgeIQRHigh) &
      \VietnamRecords \\ \tightrule

    \attrcell{San Diego, California, USA} &
      \SanDiegoDurationDays\ days &
      \SanDiegoPatients\newline(\SanDiegoMale\ M, \SanDiegoFemale\ F) &
      \SanDiegoMedianAge\newline(\SanDiegoAgeIQRLow\ -- \SanDiegoAgeIQRHigh) &
      \SanDiegoRecords \\ \tightrule

    \attrcell{New York,\newline New York, USA} &
      \NewYorkDurationDays\ days &
      \NewYorkPatients\newline(\NewYorkMale\ M, \NewYorkFemale\ F, \NewYorkOther\ O) &
      \NewYorkMedianAge\newline(\NewYorkAgeIQRLow\ -- \NewYorkAgeIQRHigh) &
      \NewYorkRecords \\ \tightrule
    
  \end{tabularx}
  \caption{\textsbf{Summary of \name clinical pilots.} M = male, F = female, O = other.}
  \label{tab:pilots}
\end{table}
% \clearpage

\linelabel{line:assembled-incrementally-start} A complete \name record is assembled from event-level messages emitted during operation (Figure~\ref{fig:elements}a). Each message is written to a collector through a dedicated write-only endpoint. At inference start, the first message records the fields available at invocation, including Header, Model instance, User identity, Target identity, and Inputs. It also creates a primary \texttt{event\_id} and, when needed, a \texttt{run\_id} for linking steps within a multi-stage workflow. Later messages reference this identifier and append schema-compliant fragments for Internal artifacts, Patient- or clinician-facing outputs, Outcomes, and User feedback. This design creates a record even when inference fails and allows Outcomes or Feedback to be added later when they become available.\linelabel{line:assembled-incrementally-end}

\section*{ICU alert monitoring with \name in Switzerland}
% BERN FIGURE
\begin{figure}[p]
  \centering
  \includegraphics[width=\textwidth]{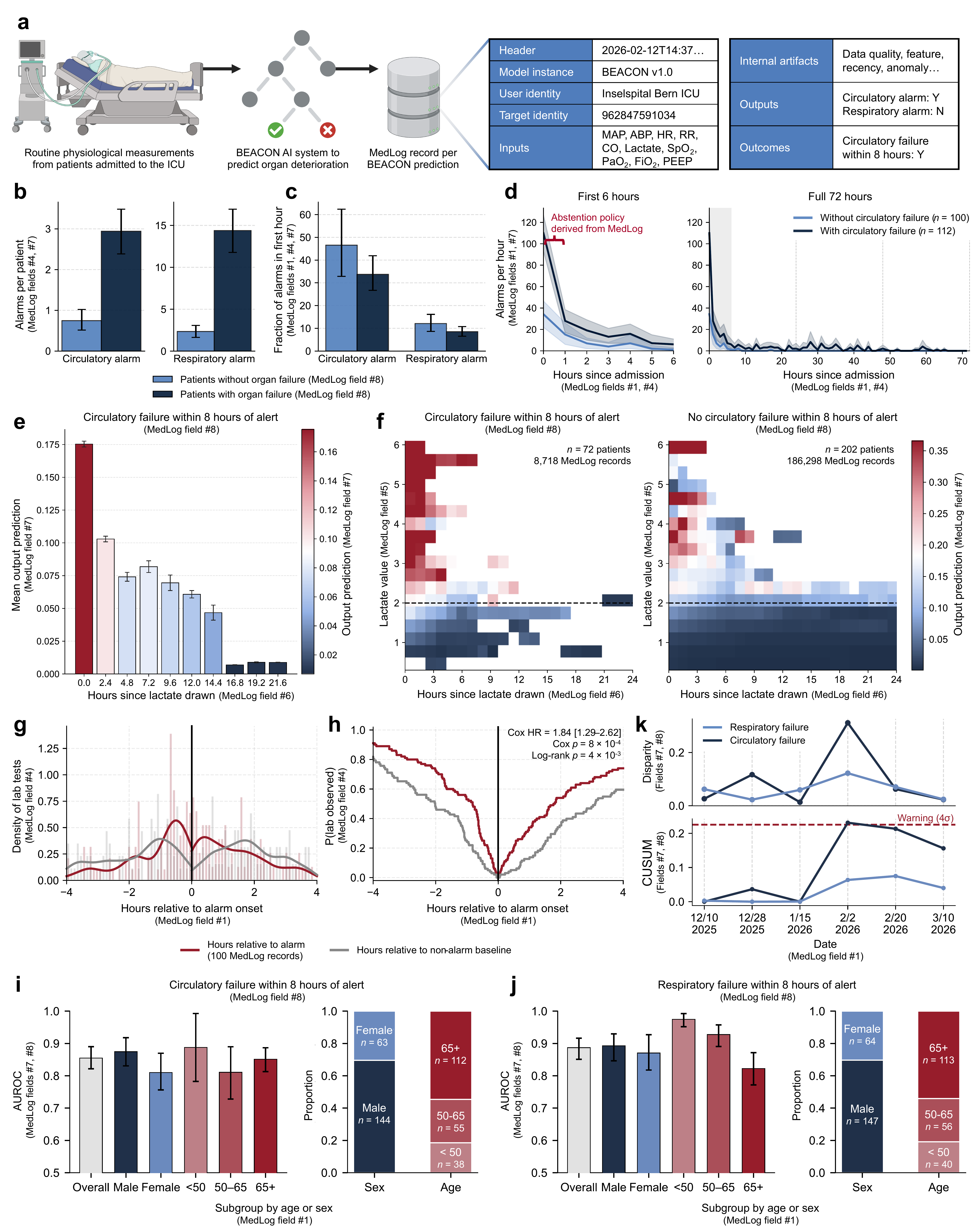}
  \caption{\textsbf{Monitoring \beacon with \name in Switzerland.} 
  \textsbf{(a)} Schematic of the \beacon AI model and its integration with \name for early warning of organ failure.
  \textsbf{(b)} Mean number of \beacon alarms per (cont.)
  }
  \label{fig:zurich}
\end{figure}

\begin{figure}[t]\ContinuedFloat
  \caption*{
  \textsbf{Figure \ref{fig:zurich} (cont.)} patient stratified by whether the patient developed a circulatory (left) or respiratory (right) organ failure event. Error bars show 95\% confidence intervals.
  \textsbf{(c)} Fraction of all \beacon alarms occurring within the first hour after ICU admission, with the same stratification.
  \textsbf{(d)} Hourly \beacon alarm rate for the first 72 hours post-admission for patients without ($n=100$) versus with ($n=112$) subsequent circulatory failure. The left panel focuses on the first six hours post-admission, where alarms concentrate; the red bracket marks the abstention policy derived from \name insights
  \textsbf{(e)} Mean $\pm$ s.e.m. \beacon output prediction by hours since the most recent arterial lactate measurement for \name records preceding a circulatory failure event within the next 8 hours. Records were grouped into 10 bins across 24 hours, and $x$-axis labels denote the lower bound of each bin. 
  \textsbf{(f)} Joint distribution of hours since the most recent arterial lactate measurement and the corresponding lactate value, with cells colored by the mean \beacon output prediction. \name records are split by whether circulatory failure occurred within 8 hours of the alarm (left, $n=72$ patients, 8,718 \name records) or not (right, $n=202$ patients, 186,298 \name records). The dashed line marks the lactate $> 2$ mmol/L circulatory failure threshold.
  \textsbf{(g)} Density of lactate laboratory test orders in the $\pm 4$-hour window around \beacon circulatory failure alarm onset (red), compared with matched non-alarm baseline time points drawn per patient (gray).
  \textsbf{(h)} Cumulative probability of a lactate laboratory test by time relative to circulatory failure alarm onset, for alarm windows (red) and matched non-alarm baseline windows (gray), log rank test on the time-to-next-lab distribution yields $p = 4\times 10^{-3}$. A Cox proportional hazards model fitted to the time-to-next-lab outcome while adjusting for time-to-last-lab yields a hazard ratio for alarms of 1.84 (95\% CI 1.29--2.62, $p = 8\times 10^{-4}$).
  \textsbf{(j)} Model performance (AUROC) for circulatory failure prediction, overall and stratified by sex and age group (left), with error bars indicating 95\% confidence intervals. The right panel shows the demographic composition of the evaluated cohort by sex and age.
  \textsbf{(k)} Model performance (AUROC) for respiratory failure prediction, overall and stratified by sex and age group (left), with error bars indicating 95\% confidence intervals. The right panel shows the demographic composition of the evaluated cohort by sex and age.
  \textsbf{(k)} Model performance disparity (AUROC) between sex groups over consecutive 18-day periods for circulatory and respiratory failure (top); and cumulative sum change-point detection statistic (CUSUM) for the same disparity (bottom), with the red dashed line showing the warning threshold.
  }
\end{figure}

% BERN TEXT
\linelabel{line:bern-pilot-start} Early recognition of physiological decline in critical care drives timely, life-saving interventions. AI-based early warning systems show promise, but performance varies across settings and transparency remains limited~\autocite{habib_fda_2023, edelson_early_2024}. We conducted a prospective three-month pilot using \name to monitor \beacon, an AI model deployed in the Intensive Care Unit (ICU) at the University Hospital of Bern that generates early warnings of circulatory and respiratory failure~\autocite{hyland_early_2020, huser_rms_2025} (Methods Section~\ref{methods:bern}, Figure~\ref{fig:zurich}a). \name records supported post-deployment analyses that revealed four insights: quantification of bedside observations, detection of failure modes not visible to end users, characterization of interactions between model outputs and end users, and variation in predictions across clinical contexts.

\linelabel{line:bern-suppression-policy-start} Following deployment, ICU clinicians reported frequent \beacon alarms immediately after admission. To quantify this pattern, we analyzed all \beacon-generated alerts across \BernPatients\ patient admissions using \name records (Figure~\ref{fig:zurich}b). Of all alarms triggered during the study period, $16.5\%$ occurred within the first hour after ICU admission (Figure~\ref{fig:zurich}c-d). This early concentration was similar across subsequent clinical trajectories: admission-hour alarm rates did not differ meaningfully between patients who later developed circulatory or respiratory failure and those who remained stable ($16.4\%$ vs. $17.4\%$ alarms per patient-hour). This pattern is consistent with the admission period, which features intense clinical activity, physiological instability, and limited data availability at a time when bedside physician attention is already high. Based on these findings, we implemented a suppression policy that withheld \beacon alerts generated within the first hour after admission from physician notification, and deployed this policy in the ICU. This example shows how \name can formalize bedside observations. \linelabel{line:bern-suppression-policy-end}

\linelabel{line:bern-shortcut-start} \name records revealed the converse, a failure mode invisible at the bedside. Even in patients who deteriorated within the subsequent eight hours, as time since the last laboratory measurement increased from under 2 hours to over 22 hours, \beacon's predicted risk fell from 0.18 to 0.01, a 95\% relative decrease (Figure~\ref{fig:zurich}e-f). This pattern indicates that the model partially uses laboratory ordering recency as a proxy for clinical concern. The effect had been anticipated but was not detected during offline development of \beacon model. \textcite{hyland_early_2020} evaluated measurement intensity bias by resampling the test set to simulate uniform measurement intervals and reported only a minor effect. However, that analysis labeled time points with missing mean arterial blood pressure or lactate measurements as ``ambiguous'' and excluded them from both training and evaluation. These are precisely the settings where laboratory data go stale and measurement recency bias takes hold. \name deployment data reveal a substantially larger real-world effect and identify a clinically important failure mode, that is, declining risk scores that clinicians may interpret as reassuring instead reflect stale data rather than improving physiology.
\linelabel{line:bern-shortcut-end}

\linelabel{line:bern-interaction-start} The \beacon AI model may also influence the clinician behavior on which it depends. To examine this possibility, we analyzed the temporal density of lactate laboratory test orders relative to circulatory failure alarm timing. Testing frequency increased from 60\% to 74\% in the 4-hour window after a \beacon alarm compared with randomly sampled control periods (hazard ratio 1.84, 95\% CI 1.29--2.62, $p = 8\times 10^{-4}$; Figure~\ref{fig:zurich}g-h), consistent with clinicians ordering confirmatory tests in response to alerts. However, this association does not establish causality. The same pattern could arise if both the alarm and the subsequent laboratory order were independently triggered by patient deterioration recognized in parallel by the algorithm and the attending physician. \linelabel{line:bern-interaction-end}

\name records also enabled fairness analysis of deployed \beacon AI model across demographic groups~\autocite{obermeyer_dissecting_2019,finlayson2021clinician}. Stratifying predictive performance by sex and age (Figure~\ref{fig:zurich}i-j) showed that AUROC was lower for female patients for both circulatory (0.81 vs.\ 0.87) and respiratory (0.87 vs.\ 0.89) failure prediction. Performance for respiratory failure prediction also declined with age, from AUROC of 0.97 in patients younger than 50 to 0.82 in those aged 65 or older. We tracked sex-based AUROC disparities for consecutive 18-day periods using a cumulative sum (CUSUM) change-point detection algorithm developed for monitoring risk prediction performance (Figure~\ref{fig:zurich}k)~\autocite{page1954continuous,feng_monitoring_2024}. For circulatory failure prediction, the sex-based AUROC disparity peaked at 0.311 in early February 2026, and the corresponding CUSUM reached 0.231, briefly exceeding the 4$\sigma$ warning threshold of 0.225 and capturing a transient spike in disparity. The subsequent decline in CUSUM suggests that this spike likely reflected small patient groups within individual time windows rather than a sustained shift. \linelabel{line:bern-fairness-end} \linelabel{line:bern-pilot-end}

\section*{Wearable tetanus monitoring with \name in Vietnam}
% VIETNAM FIGURE
\begin{figure}[!t]
    \centering
    \includegraphics[width=\textwidth]{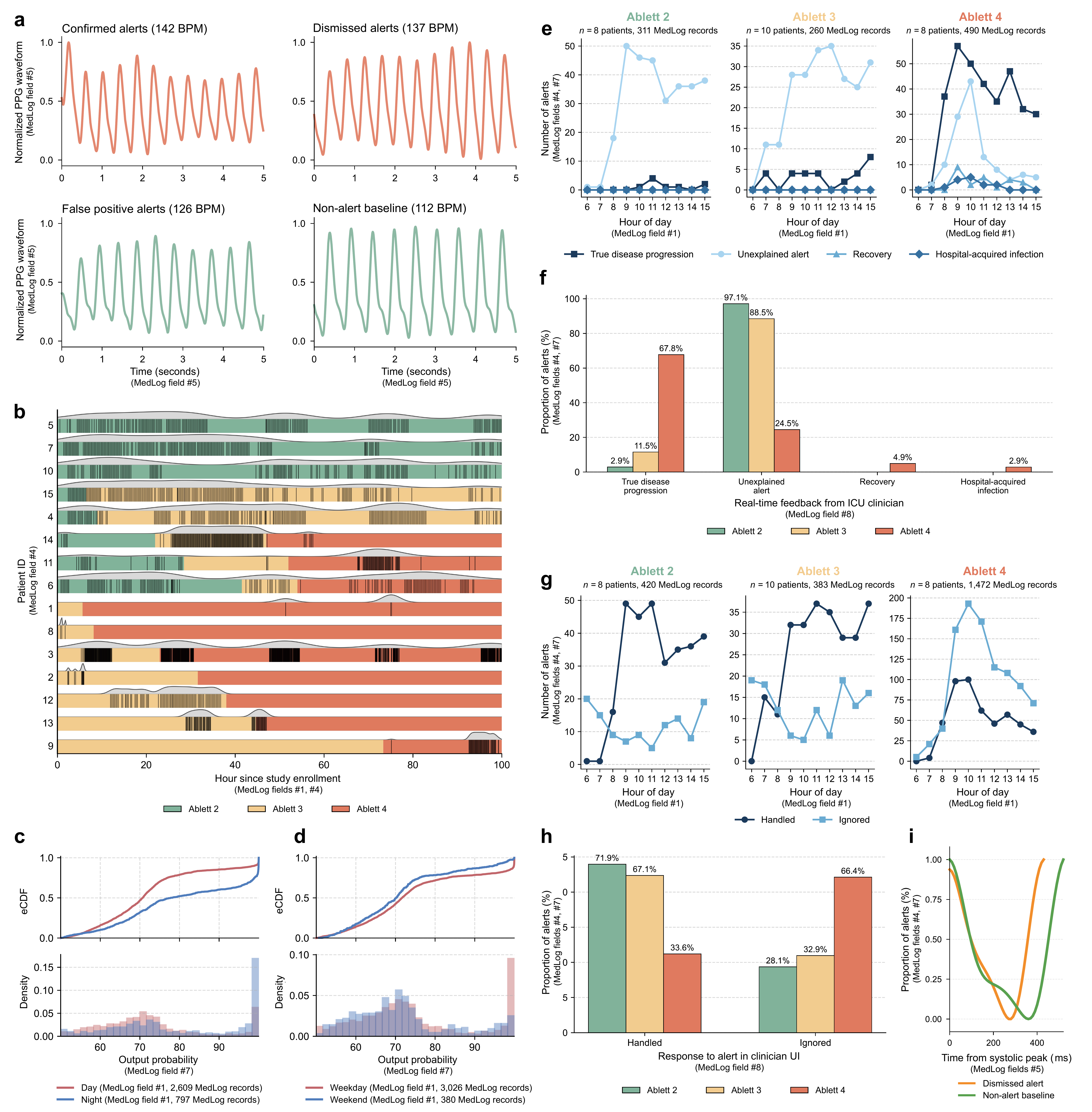}
    \caption{\textsbf{Monitoring wearable tetanus progression prediction with \name in Vietnam.}
    \textsbf{(a)} Pulse plethysmography (PPG) waveforms from tetanus patients in the clinical pilot across stages of tetanus progression.
    \textsbf{(b)} Patient trajectories and alert activity across the 15-patient cohort. Each row corresponds to a patient and shows the progression of disease severity over time based on Ablett score. Vertical ticks represent \name-recorded alert events.
    \textsbf{(c)} Empirical cumulative distribution (eCDF) functions and density histograms of \name-recorded model probabilities,
    stratified by day (07:00--18:59) versus night (19:00--06:59).
    \textsbf{(d)} Same as (c), stratified by weekday versus weekend.
    \textsbf{(e-f)} Hourly counts and within-Ablett-phase proportions of clinician-annotated reasons for each alert, stratified by Ablett phase and restricted to clinic hours (06:00--16:00).
    \textsbf{(g-h)} Hourly counts and within-Ablett-phase proportions of alerts that were handled versus ignored by the bedside clinician, stratified by Ablett phase and restricted to clinic hours (06:00--16:00).
    \textsbf{(i)} Normalized median beat morphology for dismissed alerts compared to a non-alert baseline, aligned by time from the systolic peak (ms).
    }
    \label{fig:vietnam}
\end{figure}

% VIETNAM TEXT
\linelabel{line:vietnam-pilot-start} In resource-constrained settings, where monitoring is infrequent and deterioration goes undetected, low-cost early warning systems represent the primary mechanism for triggering timely care escalation~\autocite{turner_achieving_2019}. When escalation is delayed, patients who could have been stabilized with early intervention instead progress to organ failure, septic shock, or cardiac arrest, outcomes that are both clinically devastating and resource-intensive to manage. Most AI models are developed in high-resource environments, and their performance degrades when deployed in contexts that differ from their training setting~\autocite{yang_generalizability_2024, agweyu_safety_2026}. In these high-stakes, low-resource contexts, continuous logging becomes essential to surface context-dependent failure modes and track changes in model behavior before they translate into patient harm.

\linelabel{line:vietnam-waveforms-start} We used \name to log an AI model that predicts tetanus progression from wearable physiological waveforms (Figure~\ref{fig:vietnam}a) \autocite{hai_heart_2024} in patients at the Hospital for Tropical Diseases in Ho Chi Minh City, Vietnam (Methods Section~\ref{methods:vietnam}). The model predicts deterioration defined by transitions in Ablett grade, a clinical severity scale ranging from mild rigidity (Grade 1) to autonomic instability with cardiovascular dysfunction (Grade 4). \linelabel{line:vietnam-waveforms-end} 

\linelabel{line:vietnam-trajectories-start} Using \name records, we analyzed longitudinal trajectories of patient severity defined by Ablett scores. Alerts increased during transitions to more severe disease states but remained elevated during clinical recovery (Figure~\ref{fig:vietnam}b). This pattern indicates that the model detected patient deterioration better than recovery or stable clinical states. One possible explanation is limited representation of recovery-phase waveforms in the training dataset (Methods Section~\ref{methods:vietnam}). Recovery from tetanus can span weeks to months and includes interventions such as rehabilitation, ventilator weaning, and magnesium administration, which alter physiological waveforms relative to the deterioration phase. Underrepresentation of these states in the training dataset may bias the model toward persistently elevated risk predictions. \linelabel{line:vietnam-trajectories-end}

\linelabel{line:vietnam-workflow-start} Analysis of model output probabilities recorded by \name showed variation in alert behavior across clinical workflows, with a higher proportion of high-confidence alerts at night (39.8\% alerts $\geq$ 90\% probability) than during the day (14.6\% alerts $\geq$ 90\%) (Figure~\ref{fig:vietnam}c) and on weekdays (21.4\% alerts $\geq$ 90\% probability) than on weekends (13.2\% alerts $\geq$ 90\% probability) (Figure~\ref{fig:vietnam}d). Daytime care may introduce additional physiological variability through routine nursing procedures and clinical interventions, reducing model certainty. Reduced nighttime activity may make physiological signatures of tetanus progression easier to detect. \linelabel{line:vietnam-workflow-end} 

\linelabel{line:vietnam-feedback-start} Of the \VietnamRecords\ \name records, 1,105 included real-time annotations from  clinicians on patient state and model performance. Clinician responses to alerts in the ICU user interface (UI) were recorded and classified as ``handled'' when the alert prompted interaction and ``ignored'' otherwise. All feedback and interaction traces were stored in the User feedback field of \name. As patients progressed from Ablett 2 to Ablett 4, clinicians increasingly judged alerts to reflect true disease severity rather than unexplained or false signals (Figure~\ref{fig:vietnam}e-f). In patients with Ablett 4 disease, clinicians also identified alerts associated with unrecognized recovery and hospital-acquired infections, both of which altered physiological waveforms. \linelabel{line:vietnam-feedback-end} 
\linelabel{line:vietnam-fatigue-start}In patients with Ablett 4 tetanus, alerts were more likely to be ignored than in patients with milder disease grades (Figure~\ref{fig:vietnam}g-h). This pattern may reflect alert fatigue, uncertainty about alert interpretation, or perceived redundancy when deterioration was already clinically apparent.\linelabel{line:vietnam-fatigue-end} 

\linelabel{line:vietnam-disagreement-start} To examine clinician-model disagreement, we analyzed waveforms logged with \name for a representative patient (Subject \#2) during transition to severe disease. Alerts were grouped by real-time clinician assessment and temporal proximity to transition into Ablett Grade 4: (i) confirmed alerts, validated by the clinician as Grade 4 ($n = 8$); (ii) retrospectively validated alerts, dismissed or marked uncertain at the time but occurring within 6 hours before clinical transition ($n = 5$); (iii) false positive alerts occurring more than 6 hours from a transition ($n = 20$); and (iv) a non-alert baseline representing stable Ablett Grade 3 periods ($n = 845$ windows). Visualization of raw waveforms revealed morphological signatures associated with impending deterioration (Figure~\ref{fig:vietnam}a). Confirmed and retrospectively validated alerts showed elevated heart rates (142 and 137 beats per minute [BPM], respectively) and a shallow dicrotic notch, consistent with reduced arterial compliance and increased sympathetic tone in severe tetanus. False positive alerts and baseline recordings showed lower heart rates (126 and 112 BPM) and a more pronounced dicrotic notch.

To further characterize these differences, we extracted the middle beat from the medoid window of retrospectively validated alerts and compared it with the baseline waveform (Figure~\ref{fig:vietnam}i). The baseline beat (113 BPM) showed a distinct dip caused by aortic valve closure and the reflected arterial pressure wave, whereas the dismissed beat (136 BPM) showed a shallower dicrotic notch. Because \name links waveform features to model outputs and clinical transition points, these records support analysis of physiological patterns associated with model alerts before clinical recognition of patient deterioration.\linelabel{line:vietnam-disagreement-end} \linelabel{line:vietnam-pilot-end}

\section*{Sepsis quality reporting with \name in California}
% SAN DIEGO FIGURE
\begin{figure}[ht]
  \centering
  \includegraphics[width=\textwidth]{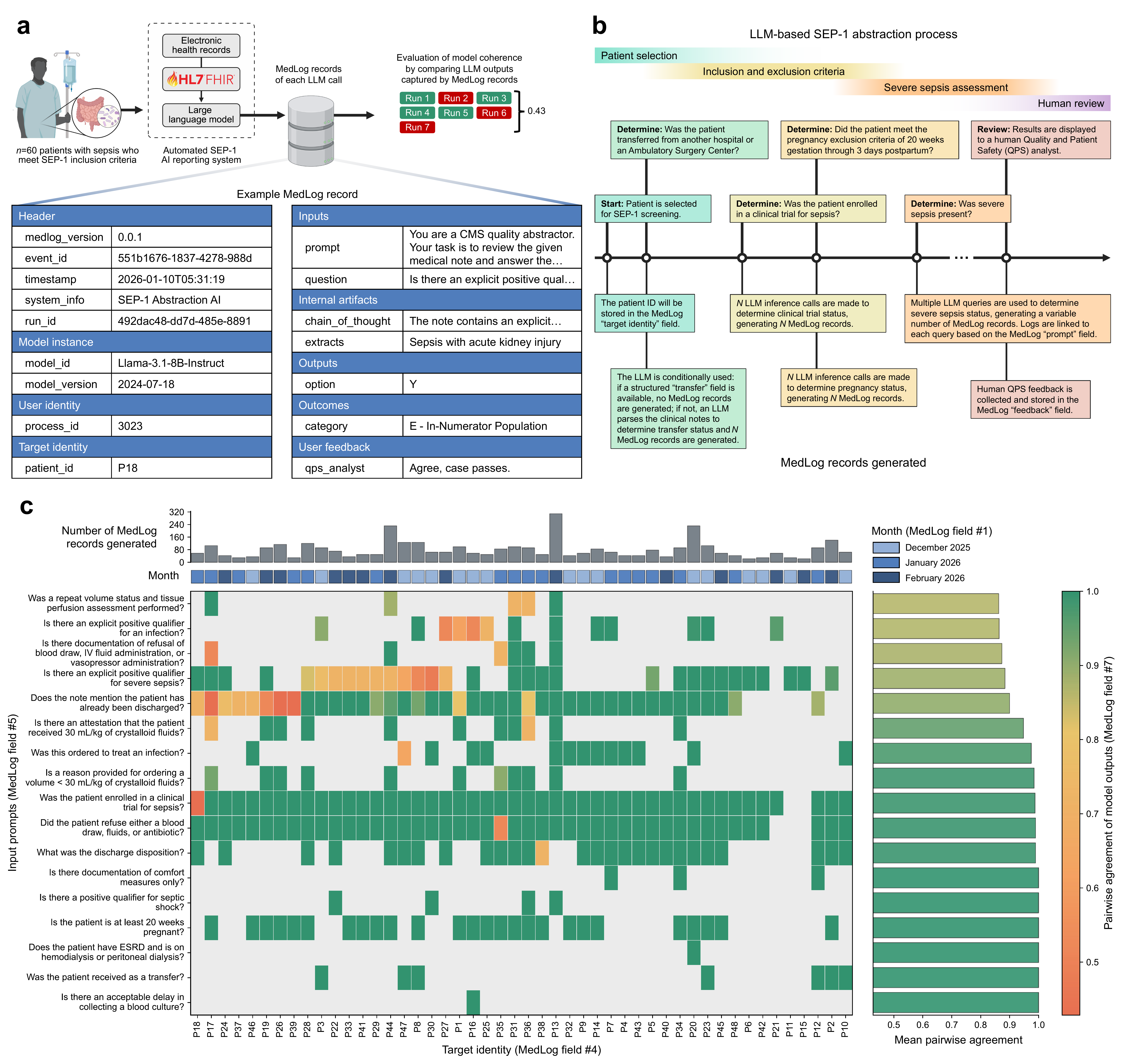}
  \caption{\textsbf{Monitoring LLM-based SEP-1 abstraction with \name in San Diego.} 
  \textsbf{(a)} Overview of the automated SEP-1 abstraction pipeline at University of California San Diego Health (UCSDH) and integration with \name. A representative de-identified \name record generated during the UCSDH prospective pilot is shown.
  \textsbf{(b)} Generation of \name records from the SEP-1 abstraction workflow. Markers indicate specific points where the LLM is queried and \name records are generated. $N$ is a hyperparameter that controls the number of LLM calls made for majority voting inference runs.
  \textsbf{(c)} Characterizing SEP-1 AI concordance across 48 patients and \SanDiegoRecords\ \name records. The heatmap shows the average pairwise agreement from best-of-$N$ sampling with majority voting for all SEP-1 questions where an LLM inference call was made. Blank cells represent questions where an LLM inference call was not required. Patients are ordered by hierarchical clustering to group cases with similar model agreement profiles.
  }
  \label{fig:ucsdh}
\end{figure}

% SAN DIEGO TEXT
\linelabel{line:ucsdh-pilot-start} The Severe Sepsis and Septic Shock Management Bundle (SEP-1) is a complex quality measure that requires an ``all-or-nothing'' 63-step abstraction process and has traditionally depended on manual chart review~\autocite{aaronson_new_2017, gesten_sep-1taking_2021}. In the United States alone, manual reporting of hospital quality measures is estimated to cost more than \$15.4 billion USD and consume 785 hours per physician each year~\autocite{casalino_us_2016}. LLMs could reduce this administrative burden, but their use for high-stakes SEP-1 quality reporting requires careful monitoring. We used \name in a prospective three-month study logging a deployed LLM for automated SEP-1 reporting at University of California San Diego Health (UCSDH) (Figure~\ref{fig:ucsdh}a-b, Methods Section~\ref{methods:sandiego})~\autocite{boussina_large_2024}.

\linelabel{line:ucsdh-coherence-start} \name logged 3,766 LLM inference events across 17 questions in the SEP-1 abstraction protocol, which were used as prompts to the LLM (median 63 \name records per patient, IQR 42--93; Table~\ref{tab:pilots}). To assess within-patient coherence, we computed average pairwise agreement across repeated LLM inferences for each patient and prompt (Figure~\ref{fig:ucsdh}c). Mean agreement was high overall at 95.4\%, with 83.6\% of patient-prompt pairs fully concordant. Consistency was highest for prompts that extracted a documented categorical value: for example, the model produced nearly identical answers when determining treatment refusal (98.9\%), clinical trial enrollment (98.8\%), and pregnancy status (100\%). Agreement was lower for prompts requiring interpretation of free-text clinical notes, for example, identification of infection events (86.4\%)  or severe sepsis (88.4\%), likely reflecting the greater complexity of these questions~\autocite{rhee_variability_2018}. Similarly, although the LLM extracted the documented discharge disposition with high consistency (98.9\%), it was less consistent when asked whether a note indicated the patient had already been discharged (90.0\%), with the LLM disagreeing at least once for 36.6\% of patients evaluated with this prompt. While discharge disposition is often explicitly documented, determining whether a discharge event has occurred requires deciding whether the movement described in a note constitutes leaving the hospital, rather than intra-hospital transfers or temporary transport for a procedure.\linelabel{line:ucsdh-coherence-end} Lastly, although our study was limited in duration, we did not observe evidence of data shift across months.\linelabel{line:ucsdh-pilot-end}

\section*{Patient attendance prediction with \name in New York}
% NEW YORK FIGURE
\begin{figure}[t]
  \centering
  \includegraphics[width=\textwidth]{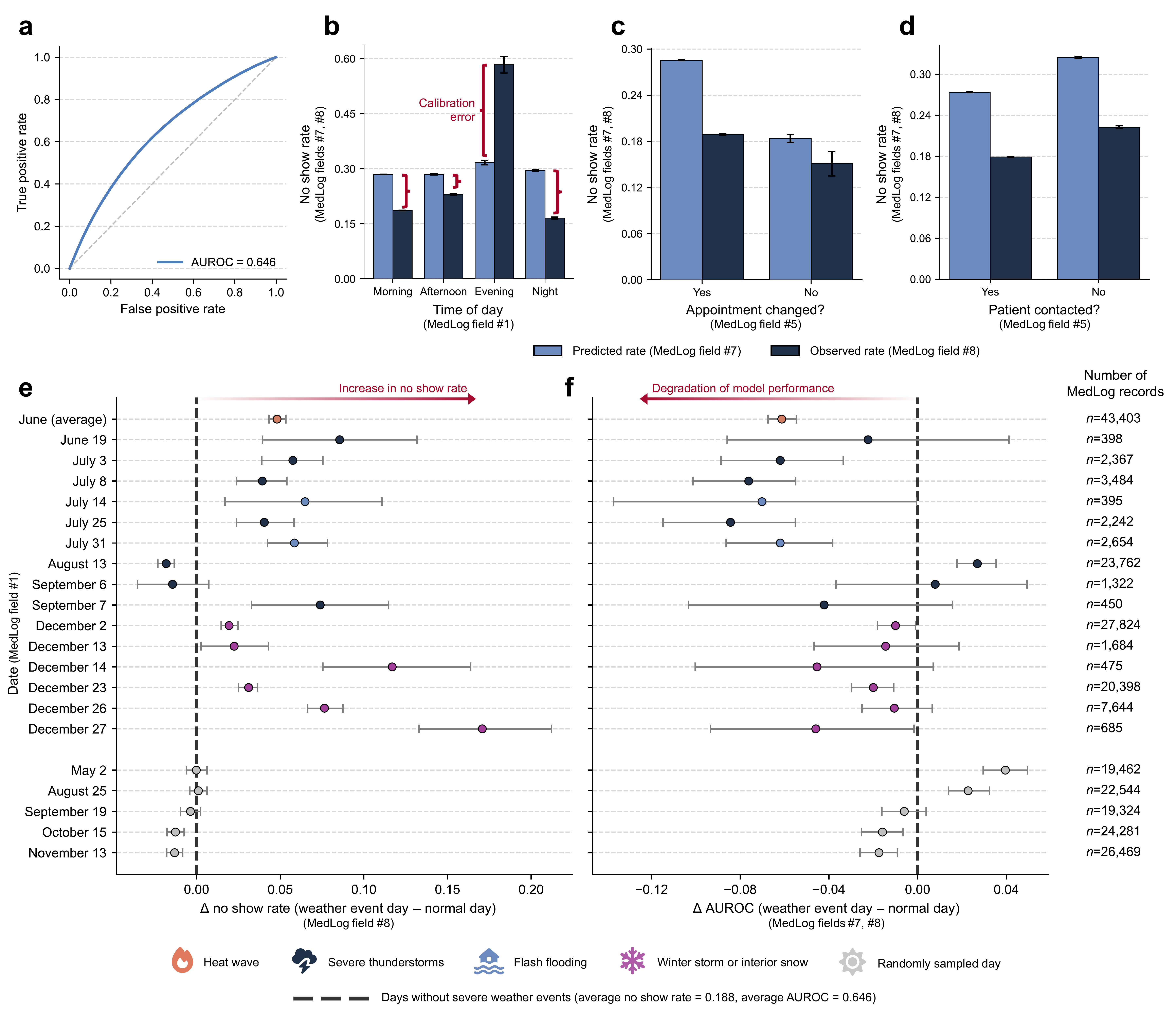}
  \caption{\textsbf{Monitoring a patient attendance prediction model with \name in New York.} 
  \textsbf{(a)} Receiver operating characteristic curve for the deployed no show prediction model across \NewYorkRecords\ \name records generated within the Mount Sinai Health System between May and December 2025.
  \textsbf{(b-d)} Mean predicted no show probability compared with the observed no show proportion, stratified by \textsbf{(b)} appointment time of day, \textsbf{(c)} whether the appointment was modified after the initial scheduling, and \textsbf{(d)} whether patient outreach calls were performed. Error bars indicate 95\% confidence intervals.
  \textsbf{(e-f)} Change in  \textsbf{(e)} observed no show rate and \textsbf{(f)} AUROC during severe weather events relative to all other clinic days. Dashed vertical lines indicate the average no show rate (0.188) and average AUROC (0.646) during baseline clinic days. Error bars indicate 95\% confidence intervals.
  }
  \label{fig:newyork}
\end{figure}

% NEW YORK TEXT
\linelabel{line:newyork-pilot-start} We applied \name to monitor the Risk of Patient No-Show Model, an AI model developed by Epic Systems that predicts the likelihood of patient attendance at scheduled appointments within the Mount Sinai Health System (Methods Section~\ref{methods:newyork})~\autocite{agovi_external_2025, patel_systematic_2026}. During an eight-month period, the model was applied to \NewYorkPatients\ patients, generating \NewYorkRecords\ \name records for analysis. This \name deployment demonstrates population-scale logging within a production scheduling system and longitudinal monitoring across heterogeneous operational settings.

\linelabel{line:newyork-variability-start} The AI model achieved an AUROC of 0.646 throughout the deployment period (Figure~\ref{fig:newyork}a). Stratification of \name records by appointment time, whether the appointment was modified after its initial scheduling, and whether a patient outreach call was documented revealed a context-dependent miscalibration between predicted and observed no show behavior. \linelabel{line:newyork-variability-time-start} For example, the model over-predicted no shows by 5.4--13.0 absolute percentage points (pp) for morning, afternoon, and night appointments, but under-predicted no shows in the evening, where the observed no show proportion (58.5\%) was nearly double the predicted probability (31.7\%), an absolute calibration gap of 26.8 pp (Figure~\ref{fig:newyork}b). Appointments scheduled in the evening, between 17:00 and 21:59, comprised only 2,122 records, or 0.07\% of all predictions. Yet, analysis of department information recorded by \name indicated evidence of dataset shift. Evening visits had a distinct service mix, with, for example, women's health appointments accounting for 42.2\% of all appointments, but only 5.9\% at other times. By contrast, the non-evening service was dominated by general (46.7\%), surgical (14.4\%), and cardiology (11.4\%) appointments, suggesting that the model, which was calibrated to this high-volume daytime population, did not capture the attendance behavior of the under-represented evening population.\linelabel{line:newyork-variability-time-end}

\linelabel{line:newyork-variability-outreach-start} The effects of administrative actions on predicted no show behavior were also miscalibrated. The model was well calibrated for appointments left unchanged after booking, with a difference of predicted versus observed no show rate of 3.2 pp. However, after an appointment had been modified by an administrative action, the model over-predicted no shows by 9.6 pp, raising predicted risk by 10.2 pp while the observed no show rate rose only 3.8 pp (Figure~\ref{fig:newyork}c). This suggests that the model overweights post-scheduling modification as a signal for future non-attendance. By contrast, for outreach calls, the model consistently over-predicted no shows by approximately 10 pp, and reproduced the effect of contacting a patient in both direction and magnitude (predicted and observed no show rate 5.1 and 4.3 pp lower, respectively, for contacted appointments), indicating a consistent pessimistic bias rather than a context-specific failure (Figure~\ref{fig:newyork}d). \linelabel{line:newyork-variability-outreach-end} \linelabel{line:newyork-variability-end}

\linelabel{line:newyork-weather-start}Finally, we examined how severe weather events affected patient attendance and model performance. Using \name records linked to severe weather events from the United States National Weather Service archive, we compared no show rates and predictive performance during weather events against all other clinic days (Figure~\ref{fig:newyork}e). Severe weather events, including heat waves, severe thunderstorms, flash flooding, and winter storms, were associated with, on average, an increase of 5.46 absolute pp in no show proportions relative to baseline clinic days. During these events, model AUROC also declined by up to 0.084 (13.0\%; Figure~\ref{fig:newyork}f). These results indicate that patient behavior during environmental disruptions differed from the conditions represented in the training data, producing context-dependent dataset shift that degraded predictive performance. \linelabel{line:newyork-weather-end} \linelabel{line:newyork-pilot-end}

\section*{Implementing \name at scale}
\label{sec:challenges}
Operational requirements for \name follow from the deployments described above, where model use must be logged while protecting patient data and integrating with clinical and computational infrastructure (Supplementary Note~\ref{si:note:further-challenges}). \name records may contain patient identifiers, clinical context, operational metadata, model outputs, and traces of clinician behavior. Deployments therefore require privacy and security safeguards used for personal health records, including compliance with HIPAA, HITECH, GDPR, and ISO/IEC 27001 \autocite{isotc_215_iso_2016, iso/iecjtc1/sc27ISOIEC270012022}, role-based access control, and audit logging.
To reduce exposure, \name can store references to clinical data and model outputs rather than raw content. This allows outputs that inform care to remain in the EHR while keeping \name separate from the legal medical record. Where regulations permit, de-identified \name records can be aggregated across sites to support post-deployment surveillance and analyses analogous to pharmacovigilance, using federated or secure computation methods \autocite{shah_nationwide_2024, yang_generalizability_2024, embi_algorithmovigilanceadvancing_2021}. Such settings require privacy-preserving data exchange, including secure multi-party computation, homomorphic encryption, or federated learning \autocite{rieke_future_2020, xia_medshare_2017}.

\linelabel{line:storage-data-start} Recording model use creates substantial storage and data management requirements. Across the four pilots, \name generated records at markedly different scales, from 3{,}406 records in tetanus monitoring and 3{,}766 in sepsis quality reporting to 223{,}840 records in ICU monitoring and \NewYorkRecords\ records from patient attendance prediction across \NewYorkPatients\ patients. These deployments illustrate how logging volume depends on workflow design, inference frequency, deployment duration, and operational scale. These requirements are consistent with broader trends in healthcare data generation, which is projected to exceed 10{,}800 exabytes annually by 2025 \autocite{banks_sizing_2020, telenti_treating_2020}, with a single hospitalization producing on the order of 150{,}000 data elements \autocite{banks_sizing_2020, esteva_guide_2019}. \name extends existing clinical data systems by recording model inputs, outputs, metadata, and optional execution traces associated with AI use. Institutions may retain all \name records for retrospective analysis or apply selective retention policies to manage storage, cost, and operational requirements. Retention strategies include full tracing during pilot studies or after major model updates, sampling or risk-triggered tracing during steady-state deployment, and tiered retention with long-term summaries and shorter-lived detailed artifacts.\linelabel{line:storage-data-end} 

\name can be deployed within a health system without coordination with external institutions or vendors. Cross-site analysis, however, requires shared interfaces across EHR vendors, AI vendors, and health systems. \name can be integrated into existing systems at API gateways, LLM proxies, or similar interfaces that intercept model calls and emit \name-compliant records. The same approach can wrap agent frameworks and tool calls to record inputs, retrieved context, outputs, and uncertainty estimates. This design supports incremental adoption without modifying the underlying models.

Our real-world implementations show that \name can be implemented with existing standards and tooling. In the ICU and tetanus deployments, \name records tracked repeated model calls, clinical context, and outcomes over time. In the sepsis reporting deployment, \name recorded repeated LLM calls within a multi-step abstraction workflow. Existing standards can represent these patterns without requiring a new software stack. The W3C PROV model provides one such framework for computational provenance, where \name records map to \texttt{prov:Entity}, model invocations to \texttt{prov:Activity}, and users or models to \texttt{prov:Agent} \autocite{khalid_belhajjame_prov-o_2013}. OpenTelemetry can collect and transport these events, and Fast Healthcare Interoperability Resources (FHIR) can anchor \name records to standardized clinical entities such as AuditEvent, Patient, Condition, Observation, Practitioner, and PractitionerRole \autocite{hl7_fhir_2023}.

\name can be implemented in low- and lower-middle-income settings without requiring full EHR infrastructure or continuous connectivity. A minimal implementation can record only a small set of fields, such as Header, Model instance, and Outputs, and add further fields as local capacity grows. When network access is intermittent, local write-behind caching can store records on device and synchronize them once connectivity returns. Where EHR systems are unavailable, \name records can instead link to encounter-level metadata such as time, location, and department, with optional FHIR integration when available. \name can also connect to platforms such as OpenMRS \autocite{mamlin_cooking_2006} and DHIS2 \autocite{dehnavieh_district_2019}. 

Implementing \name requires more than technical integration. It also requires clear operational ownership, defined review workflows, and rules for access to detailed records \autocite{kadakia_modernizing_2021}. Health systems can keep full \name records under local control while using aggregated analyses for quality improvement, benchmarking, or surveillance. Clinician involvement is important in deciding which events to review, how to interpret logged outputs, and how to act on recurring failure modes. 

Detailed \name records can expose AI systems to security risks, including membership inference and model extraction attacks \autocite{shokri_membership_2017, mattern_membership_2023, zhao_systematic_2025, sha_prompt_2024, zhang_extracting_2024}. Reducing these risks requires safeguards and governance mechanisms that limit reverse engineering \autocite{he_cater_2022, li_llm-pbe_2024}. Data ownership and provenance metadata within \name records can further clarify responsibilities among health systems, developers, and regulators, and aligning incentives

\section*{Discussion} \label{sec:future}
\name creates an event-level record of how AI models are used in medicine, capturing inputs, outputs, intermediate artifacts, outcomes, and user feedback for each model invocation. Across the deployments presented here, these records make model behavior visible within real-world use in ways that pre-deployment evaluation cannot. They surface failure modes, quantify performance variation across medical contexts, and link model outputs to subsequent actions in patient care (Figure~\ref{fig:overview}).
We discuss other uses of \name in Supplementary Note~\ref{si:note:further-discussion}, and limitations in Supplementary Note~\ref{si:note:limitations}.

\begin{figure}[!t]
  \centering
  \includegraphics[width=\textwidth]{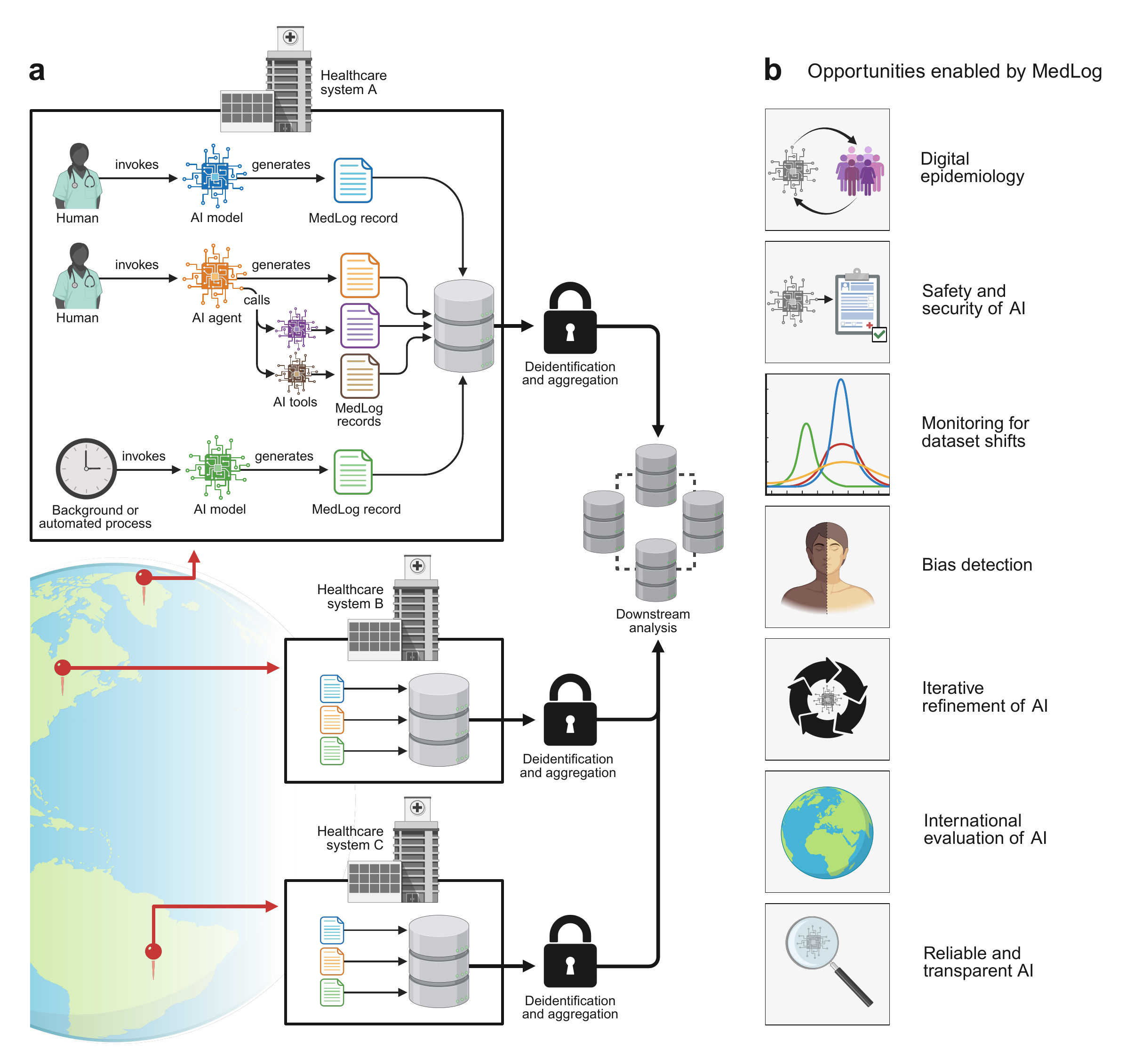}
    \caption{\textsbf{Implementing and using \name.}
    \textsbf{(a)} \name can be implemented within institutions, across health systems, and across regions. De-identified records can be aggregated for local and cross-site analysis. \textsbf{(b)} \name records support analysis of human-AI interactions, safety and security monitoring, detection of dataset shift and bias, model refinement, evaluation across settings, and post-deployment surveillance of medical AI.}
  \label{fig:overview}
\end{figure}

\xhdr{Epidemiology of medical AI}
Continuous logging creates a data layer that provides systematic event-level records of how AI models are used in medicine. \name captures what information a model receives, what it produces, and what follows, linking inputs, outputs, and downstream events at the level of individual model invocations. These records make AI behavior measurable in deployment, supporting analysis of how models shape decisions, where performance improves \autocite{mcduff_towards_2025} or degrades \autocite{budzyn_endoscopist_2025}, how behavior varies across populations and settings, and how workflows change when AI is introduced. In the Bern ICU deployment, \name records showed that alerts concentrated in the first hour after admission and preceded increased laboratory testing, revealing a workflow-dependent pattern in model behavior and a measurable association between model outputs and subsequent physician actions. These analyses can inform quality improvement, deployment decisions, and health policy at scale.

\xhdr{Real-time surveillance of medical AI} \linelabel{line:real-time-surveillance-start}
Drawing on approaches established in cybersecurity and distributed computing, \name logs can be analyzed in real time to detect anomalous model behavior, identify unexpected output patterns, and trigger human review before failures propagate into patient care \autocite{zhu_loghub_2023, qi_loggpt_2023, karlsen_benchmarking_2024}. Supervisor agents or automated pipelines can continuously process \name records, flagging deviations from expected performance distributions and escalating cases that warrant expert investigation, making organization-wide oversight operationally feasible without proportional increases in human review burden. This real-time layer also generates the audit trail that regulatory frameworks require \autocite{raji_closing_2020, liu_medical_2022}, including recent guidance from the U.S. Food and Drug Administration and related initiatives \autocite{warraich_fda_2024, center_for_veterinary_medicine_considerations_2025, center_for_devices_and_radiological_health_artificial_2025, shah_nationwide_2024}. In the New York deployment, continuous monitoring across \NewYorkRecords\ records detected workflow-dependent calibration differences and performance degradation during severe weather events, failures that pre-deployment evaluation did not reveal and that only emerged through longitudinal monitoring at population scale.\linelabel{line:real-time-surveillance-end}

\xhdr{Detecting dataset and model shifts}
\name records detect changes in model behavior when patient populations, clinical workflows, or deployment environments diverge from those represented in training data (Supplementary Note~\ref{si:note:casestudy}, Supplementary Figure~\ref{si:fig:casestudy}) \autocite{subbaswamy_development_2020, finlayson_clinician_2021}. In the Vietnam deployment, alert behavior differed between day and night and between weekdays and weekends, showing that model outputs varied with the care environment. In the New York deployment, severe weather events drove higher no-show rates and lower AUROC, demonstrating that operational disruptions induce measurable dataset shift in real-world prediction tasks. Event-level records distinguish shifts in input data from shifts in how models are used, including changes in monitoring intensity, workflow timing, patient behavior, and operational context~\cite{li2026scaling}. Hypothesis testing approaches applied to \name records detect global and subgroup-level shifts that aggregate monitoring misses \autocite{koch_distribution_2024, schrouff_diagnosing_2022, koch_hidden_2022}. \name also tracks performance changes after model updates or retraining, including degradation introduced by corrupted or manipulated training data \autocite{alber_medical_2025}.

\xhdr{Monitoring bias}
Stratifying \name records by patient attributes reveals differential model performance across demographic groups \autocite{seyyed-kalantari_underdiagnosis_2021, daneshjou_disparities_2022, liu_translational_2023, xiong_how_2025}. In the Bern deployment, \name records showed lower predictive performance for female patients in circulatory and respiratory failure prediction, and for older patients in respiratory failure prediction. The same approach applies to LLM-based systems, and in the San Diego deployment, \name records tested whether agreement rates, error patterns, and abstraction quality differed across patient groups and care contexts. Combined with slice discovery methods, event-level records identify subpopulations with distinct AI performance profiles \autocite{eyuboglu_domino_2022}, surfacing disparities that aggregate metrics obscure and directing human review where it is most needed \autocite{liu_translational_2023, de_kanter_preventing_2023}.

\xhdr{Using \name data to improve medical AI}
\name records error cases, uncertainty estimates, and user feedback that together drive targeted model review and updates \autocite{budd_survey_2021}. These records guide data selection for model retraining, support continual learning as new observations accumulate, and inform adjustments to system behavior based on observed use patterns \autocite{bengio_curriculum_2009, liu_competence-based_2023, rui_improving_2025, zheng_lifelong_2025, gao_survey_2025}. Information captured by \name yields evaluation datasets grounded in real-world use, producing benchmarks that better reflect deployment conditions than those constructed from held-out training data \autocite{lin_wildbench_2024}.

\xhdr{International evaluation of AI}
\name records aggregated in de-identified form reveal how model performance vary across geographic and economic settings. The deployments presented here span ICU and outpatient workflows, infectious disease monitoring and sepsis prediction, predictive models and LLM-based systems, and both resource-constrained and high-resource environments. Low- and middle-income countries remain severely underrepresented in medical AI research \autocite{yang_disparities_2024, han_randomised_2024}, even as real-world deployment in these settings accelerates \autocite{mateen_trials_2025, korom_ai-based_2025}. Systematic logging with \name measures model generalizability across settings, surfaces context-specific limitations \autocite{yang_generalizability_2024}, and produces the comparative deployment data needed to determine whether AI narrows or widens performance gaps between resource-rich and resource-constrained health systems.

\xhdr{Transparent use of medical AI}
\name records document AI-generated outputs within medical and administrative workflows, creating a traceable account of how AI participated in each encounter. Where patient-accessible health data exports exist, such as through the ``Blue Button'' initiative \autocite{assistant_secretary_for_technology_policy_blue_2022, mandl_push_2020}, \name records can be incorporated directly, giving patients, caregivers, and downstream systems visibility into the AI interactions that shaped their care \autocite{khasentino2025personal}.

\section*{Conclusion} \label{sec:conclusion}
AI is increasingly used in medicine, but its deployment is rarely recorded in ways that support systematic evaluation. Across the deployments presented here, event-level logging revealed model behavior within real-world workflows that pre-deployment evaluation did not anticipate, including workflow-dependent performance variation, differential performance across patient groups, measurable associations between model outputs and physician behavior, and performance degradation driven by severe weather events in patient attendance prediction. \name establishes a standardized event-level record of medical AI use, capturing the information needed to monitor safety, detect bias and dataset shift, satisfy audit requirements, and guide model refinement. As AI becomes embedded in medicine, logging transforms deployment from an unobserved process into a source of evidence for improving models, informing policy, and protecting patients.

\clearpage

\section*{Methods} \label{sec:methods}
\setcounter{section}{0}

For all deployments, we used the \name specification available at \url{https://medlogprotocol.ai} as the reference implementation. The specification uses an  HTTP REST interface, but \name is transport-agnostic and can be implemented across different healthcare environments. We adapted \name in each deployment setting to match local infrastructure, logging requirements, analysis needs, and resource constraints. Future work and community consensus are needed to define interoperability standards for \name.

Because each patient contributed many \name records, records represent repeated measurements nested within patients rather than independent observations. To account for this within-patient correlation, confidence intervals were obtained by patient-level cluster bootstrapping and regression models were clustered at the patient level.

\section{Bern, Switzerland: ICU deterioration monitoring}\label{methods:bern}

\xhdr{Background on AI model development} \name was used to log \underline{B}ackground \underline{E}arly \underline{A}wareness for \underline{C}ritical deteriorati\underline{ON} (\beacon), an AI system that analyzes routinely collected bedside and laboratory data to predict impending circulatory ~\autocite{hyland_early_2020} or respiratory failure~\autocite{huser_rms_2025}. The circulatory model operates on 25 clinical variables, expanding to almost 600 temporal statistical features provided to the boosted tree model. The respiratory model operates on 18 clinical variables, expanding to over 300 temporal statistical features provided to the algorithm. Circulatory failure was defined as lactate $> 2$ mmol/L together with mean arterial pressure $\leq 65$ mmHg or receipt of vasopressive or inotrope drugs, and respiratory failure was defined as a ratio of partial pressure of arterial oxygen to fraction of inspired oxygen (PaO\textsubscript{2}/FiO\textsubscript{2}) $< 150$ mmHg.

 The two models were trained on the HiRID-II~\autocite{huser_rms_2025} dataset, which comprises over $55,000$ patients admitted to the interdisciplinary adult intensive care unit at the University Hospital of Bern in Switzerland from 2008 until 2019. A previous version of this dataset, HiRID-I, is publicly available~\autocite{hirid-i-faltys2021} and comprises $34,000$ patients admitted until 2016. The models were evaluated on a temporal data split by holding out the last year of data, from mid-2018 until the end of data collection in 2019, to reproduce performances reported by \textcite{hyland_early_2020} and \textcite{huser_rms_2025} for circulatory and respiratory failure, respectively. In 2024, the hospital transitioned to a new data management system, necessitating a revised data mapping. Given that training data included only patients admitted up to 2018, to ensure robustness to temporal shift and data infrastructure changes, the models were first evaluated on a retrospectively extracted cohort from the new data management system, then silently deployed for a brief period before being released for the interventional study described below.

\xhdr{Deployment and \name integration} We embedded \name into a single-center, stratified, cluster-randomized multiple-period crossover (CRXO) trial (\href{https://clinicaltrials.gov/study/NCT07119411}{NCT07119411}) conducted within the Blue and Yellow adult sub-units of the ICU at the University Hospital of Bern. The aim of this study was to evaluate whether providing \beacon alerts alongside routine clinical ICU monitoring improves unit-level outcomes such as 28-day mortality and the severity of organ failures. Our \name pilot includes three months of prospectively collected data from the \beacon trial cohort. Eligible patients were emergency ICU admissions, excluding those with declined general research consent, younger than 18 years, and those admitted solely for end-of-life care or organ donation, or evaluation of such. \name was integrated into the trial infrastructure to monitor every \beacon prediction and alert across 212 patient admissions in the interventional arm of the study.

The BEACON model was run on securely hosted virtual machines directly inside the hospital infrastructure. A data coding proxy mediated between real-time production data feeds and the machine learning backend, as well as the research database and both coded and non-coded graphical user interfaces. Live data updates were retrieved at minute-level intervals from the hospital's Epic system using Prefect workflow orchestration. Every five minutes, the data processing pipeline was triggered to compute partial updates to patient state representations using newly available data, and risk scores were recomputed. At this stage, \name was integrated to capture the full inference context, including model inputs, outputs, and post hoc analytes such as variable importance scores. To efficiently integrate \name into the production system with minimal changes, we implemented a reference-based approach, in which each \name record contained pointers to corresponding cells in the underlying database schema.

\name records were stored as a Parquet data table, with one row for each five-minute prediction interval per patient across the full ICU stay. Data were exported from the coded production research database, encrypted, and transferred via secure channels to \href{https://sis.id.ethz.ch/services/sensitiveresearchdata/}{Leonhard Med}, a trusted research environment at ETH Zurich~\autocite{okoniewski_leonhard_2024}. The columns of the table corresponded to \name fields, including \beacon-predicted output probabilities for circulatory and respiratory failure, the binary alarm indicators, currently observed data, historical model input features, and for every input laboratory analyte, both the time since the last measurement and the time until the next measurement, which were retrospectively computed. Alarm events are derived from prediction scores using the policy introduced by \textcite{hyland_early_2020}. A target deployment threshold is fixed, set here to 80\% event recall on the test cohort of the HiRID-II dataset. Whenever a prediction score exceeds this threshold, an alarm is issued, after which further alarms are suppressed for a predetermined silencing period. To further reduce alarm burden in deployment, the policy was extended with an exponential back-off that progressively lengthens the silencing period for patients in prolonged risk states.

\xhdr{Analysis of \name records} We analyzed \BernRecords\ \name records generated from \BernPatients\ patients. For alarm burden analysis (Figure~\ref{fig:zurich}b-d), 95\% confidence intervals were obtained by a patient-level cluster bootstrap with 500 resamples. For laboratory recency analysis (Figure~\ref{fig:zurich}e-f), mean circulatory failure predictions were binned by time since the last lactate measurement and stratified by whether circulatory failure occurred within 8 hours. Joint heatmaps of predicted risk over lactate value and measurement recency reveal how predictions degrade as laboratory data become stale, separately for time steps with and without impending organ failure. For alert influence analysis (Figure~\ref{fig:zurich}g-h), we estimated the time from each circulatory failure alarm onset to the next arterial lactate draw using Kaplan-Meier cumulative incidence curves within a $\pm$4-hour window, censoring at the observation window boundary or the next alarm. Only alarm onsets preceded by at least 8 hours without a prior alarm were included to assess the effect of a novel alert. A matched baseline distribution was constructed by sampling non-alarm time points from the same patient within a 72-hour window around the alarm and computing a Kaplan-Meier curve under prior censoring. A two-sided log-rank test was used to assess the statistical significance of the difference between the two curves. A two-sided Cox proportional hazards model, clustered on patient, estimated the hazard ratio for lab acquisition after alarm versus baseline, adjusted for time since the last laboratory measurement.

For fairness analyses (Figure~\ref{fig:zurich}i-k), 95\% confidence intervals were obtained by patient-level bootstrapping with 500 samples. We evaluated predictive performance by the area under the receiver operating characteristic curve (AUROC) stratified by sex and age group. Performance disparity is defined as the difference in AUROC between the largest and smallest performance across subgroups. Change-point detection was done using the one-sided cumulative sum (CUSUM) algorithm~\autocite{page1954continuous}, with the baseline mean and standard deviation $\sigma_0$ estimated using the first $2$ time windows (\ie, one-third of available windows), an allowance parameter of $0.5 \sigma_0$, and a warning level of $4\sigma_0$.

\section{Ho Chi Minh City, Vietnam: Wearable tetanus monitoring}\label{methods:vietnam}

\xhdr{Background on AI model development} \name was used to log a pulse plethysmography (PPG)-based AI model to predict tetanus severity~\autocite{hai_heart_2024, karolcik_towards_2024, lyle_artificial_2025}. The AI model consisted of a pre-trained vision Transformer base and a binary classification head. Masked patch prediction was used for self-supervised representation learning on PPG spectrograms from illnesses involving infection and systemic instability, collected mainly from the Hospital for Tropical Diseases (HTD) in Vietnam. This training strategy was designed to encode transferable representations of infection severity from patients with different diseases treated in similar healthcare environments, improving feature extraction and stabilizing downstream performance on smaller, task-specific datasets. The model was then fine-tuned on three tetanus datasets from different clinical research studies to classify the most severe form of tetanus from all other clinical stages. Note that two of the three training datasets, comprising over 70\% of the patients, were originally curated to capture broad health state changes between the first and fifth days of hospitalization, which is the approximate time required to reach peak tetanus severity. PPG waveforms were not collected after this point; as a result, the recovery state was underrepresented in the training data. 

\xhdr{Deployment and \name integration} We integrated \name into a pilot study of the AI model at the Oxford University Clinical Research Unit (OUCRU), in the Intensive Care Unit at HTD in Ho Chi Minh City, Vietnam. Enrolled patients underwent continuous monitoring with a medical-grade pulse oximeter (SmartCare; Shanghai Berry Electronic Tech Co., Ltd., China). The device collected PPG waveforms, heart rate, and oxygen saturation, and transmitted the data via Bluetooth to a bedside tablet acting as a gateway. The waveform data were subsequently relayed to a local laptop server (Intel Core i7-1225U 1.70 GHz, 16 GB RAM) that hosted the AI model for inference. PPG segments of 3-5 minutes in duration were provided as input to both the AI model and a data quality control pipeline~\autocite{le_vital_sqi_2022}. The system generated two types of alerts: (i) severity alerts and (ii) data quality alerts (\eg, oximeter probe displacement and low perfusion index). The AI-based system directly notified ICU clinicians when increases in tetanus severity were predicted. Alerts were delivered via three channels: email, a study team Microsoft Teams channel, and on-device notifications on the laptop. Data quality alerts were intended for nursing staff, while severity alerts were intended for doctors. For this pilot study, the alert was directed to a specific study clinician who was not involved in the patient's care but evaluated the patient's status after receiving the alert. An alert was only raised to the study clinician, and a corresponding \name record generated, when the model's predictive probability crossed a 50\% threshold.

\xhdr{Analysis of \name records} We analyzed \VietnamRecords\ \name records generated from \VietnamPatients\ patients. For patient trajectory analysis (Figure~\ref{fig:vietnam}b), each alert was expressed in hours since participant enrollment and restricted to the first 100 hours. For the clinician response analyses (Figure~\ref{fig:vietnam}e--h), each alert was assigned to one of three Ablett grades (Grades 2-4), and only alerts occurring during daytime hours (06:00--16:00) where study clinicians were available to respond to or annotate alerts were retained. Clinician feedback on model alerts was normalized and mapped to four canonical categories: true disease progression prediction, unexplained alert, recovery phase, and hospital-acquired infection. Alerts were also mapped against pre-defined endpoints of severity recognition by staff providing routine clinical care in the ICU who were unaware of alerts.

For the patient-specific waveform analysis, waveforms were first preprocessed to ensure standardized comparisons. The PPG signal was bandpass filtered using a 3rd-order Butterworth filter with cutoff frequencies of 0.5 Hz and 8 Hz. The lower cutoff removes baseline drift and respiratory frequency components. The upper cutoff removes high-frequency noise. The filter was applied using zero-phase forward-backward filtering to avoid distortion. To allow comparison of waveform morphology across windows with different absolute amplitudes, the filtered signal was then min-max normalized to the range [0, 1]. Systolic peaks were detected using local maximum detection with minimum inter-peak distance and prominence thresholds. The minimum inter-peak distance was set to 0.33 seconds, corresponding to a maximum heart rate of approximately 180 BPM, with a minimum prominence of 0.4 standard deviations. Windows with fewer than 4 detected peaks were excluded from downstream analyses. Heart rate was estimated from the detected peaks by computing the mean inter-beat interval, filtering out physiologically implausible intervals, and converting to beats per minute. 

Alerts were grouped into four categories: (i) confirmed alerts ($n = 8$); (ii) retrospectively validated alerts ($n = 5$); (iii) false positive alerts, occurring more than 6 hours before or after the transition to Ablett Grade 4, excluding those alerts labeled by the ICU study doctor as ``Noised data'' ($n = 24$ alerts, of which $20$ had usable PPG data logged via \name); and (iv) a non-alert baseline, comprised of 1-minute at Ablett Grade sampled from multiple PPG recording periods across six days ($n = 845$ windows). PPG data were segmented into 30-second windows. To identify representative pulse morphologies for each group, we employed a geometric medoid approach. Within each 30-second window, individual cardiac cycles were extracted using peak-to-peak detection. Each beat was normalized, resampled to 80 samples, and then aggregated into a single five-second median beat representing that window. To select representative windows, a pairwise correlation distance matrix was computed across all windows for each group; then, the medoid was selected. Raw PPG data from these medoid windows were bandpass filtered to remove noise and visualized (Figure~\ref{fig:vietnam}a). To quantify the differences in pulse contour between clinician-model disagreement and baseline, we performed a single-beat overlay analysis. The central beat was extracted from the previously identified medoid windows and plotted in milliseconds (Figure~\ref{fig:vietnam}i).

\section{San Diego, California: Sepsis quality reporting}\label{methods:sandiego}

\xhdr{Background on AI model development} \name was used to log an automated SEP-1 reporting system at UCSDH~\autocite{boussina_large_2024}. This system was implemented into routine Quality and Patient Safety (QPS) workflows for SEP-1 reporting to the Centers for Medicare \& Medicaid Services (CMS) in June 2025. It retrieves structured and unstructured data via Fast Healthcare Interoperability Resources and queries Llama-3.1-8B-Instruct with chain-of-thought prompting, best-of-$N$ sampling with majority voting, and retrieval-augmented generation over patient notes to produce a complete SEP-1 abstraction. The model was hosted within a secure HIPAA-compliant Amazon Web Services (AWS) virtual private cloud on a single virtual machine with consumer-grade hardware. A \name record was generated for each inference call, and all records from the batch were preserved as a file in S3 cloud storage.

\xhdr{Deployment and \name integration} We conducted a three-month prospective pilot of \name on all UCSDH patients included in standard SEP-1 reporting, comprising \SanDiegoPatients\ patients (Table~\ref{tab:pilots}). Each time Llama-3.1 was invoked within the SEP-1 AI system, a \name record was created consisting of nine core fields: header, model, user, target, inputs, artifacts, outputs, outcomes, and feedback, providing a structured and consistent record of model activity (Figure~\ref{fig:ucsdh}c).

\xhdr{Analysis of \name records} The system utilized the LLM for 48 of the included patients, generating a total of \SanDiegoRecords\ \name records, which were analyzed for average pairwise agreement as a measure of model coherence. For every unique patient and SEP-1 question, the system could perform multiple runs, where each run corresponded to a single execution of the model on a particular section of the patient's notes. Within each run, Llama-3.1 generated $N$ samples; for our pilot, $N = 7$. We quantified within-run agreement as the fraction of unordered answer pairs that shared the same value. We then averaged these run-level agreement values across runs to obtain one concordance score per patient-question pair. Patients were ordered by Ward linkage hierarchical clustering on the Euclidean distance between their question-level agreement vectors, and SEP-1 questions were sorted in ascending order of their mean agreement averaged across patients.

\section{New York, New York: Patient attendance prediction}\label{methods:newyork}

\xhdr{Background on AI model development} \name was used to log the Risk of Patient No-Show Model (Epic Systems, Version 2), a supervised machine learning model to predict the likelihood of a patient not presenting for a scheduled appointment~\autocite{agovi_external_2025, patel_systematic_2026}. The model is implemented as a gradient-boosted decision tree ensemble using LightGBM, and was developed using data from over 7 million appointments collected across two healthcare systems between 2014 and 2017. Input features comprise 22 variables capturing appointment characteristics and patient history. Hyperparameter optimization was performed using grid search over key parameters, including the number of samples per leaf, the maximum tree depth, the maximum number of features considered at each split, and the minimum number of samples required to split an internal node. 

\xhdr{Deployment and \name integration} The no-show prediction model was deployed within the Mount Sinai Health System using Epic's Nebula infrastructure for both training and inference. The model artifact was packaged as a Python archive and deployed through a GitLab CI/CD pipeline. At runtime, Epic Chronicles extracts relevant patient and appointment features and submits them as a JSON request to the Nebula service, which returns a predicted no-show probability. This prediction is stored within the EPT contact record and surfaced to users through multiple Epic interfaces: the Appointment Desk Activity, the Storyboard, and optionally as a column in the Multi Provider Schedule.

The model runs automatically at the time of appointment scheduling, and also nightly beginning one week prior to the scheduled appointment start time. The no-show prediction is primarily used by appointment schedulers, including both the central scheduling team (Access Center) and staff within individual clinics and practices. Risk scores are visible within Epic at the time of scheduling and during schedule review workflows, including in the Department Appointments Report (DAR), which provides a schedule view similar to the Appointment Desk. When a patient is identified as high risk for non-attendance, staff may use this information to inform operational decisions, such as strategic overbooking, prioritizing follow-up scheduling, initiating targeted outreach, or documenting repeated no-show behavior. Notification and downstream clinical response to elevated risk scores vary by department and local workflow practices.

\xhdr{Analysis of \name records} We analyzed \NewYorkRecords\ \name records generated from \NewYorkPatients\ patients between May 2025 and December 2025, corresponding to the full period during which Version 2 of the model was active in production during 2025. To exclude cancellations, appointments were filtered to include only those that had at least one model prediction made less than 24 hours before the scheduled appointment. For each included appointment, the final prediction generated before the scheduled appointment time was used.

Model performance was summarized by area under the curve receiver operating characteristic curve (AUROC), computed across all retained records (Figure~\ref{fig:newyork}a). To characterize calibration across operational contexts, \name records were stratified by appointment time of day (morning: 04:00--11:59, afternoon: 12:00--16:59, evening: 17:00--21:59, and night: 22:00--03:59; Figure~\ref{fig:newyork}b), by whether the appointment was modified after initial scheduling (Figure~\ref{fig:newyork}c), and by whether a patient outreach call was documented (Figure~\ref{fig:newyork}d). Within each subgroup, the mean predicted no show probability was compared against the observed no show proportion. For all estimates, 95\% confidence intervals were obtained by a patient-level cluster bootstrap with 500 resamples.

To evaluate performance degradation due to external weather events (Figure~\ref{fig:newyork}e-f), \name records were linked to severe weather events between May and December 2025, retrieved from the New York Significant Weather Events Archive from the U.S. National Weather Service. Weather events were grouped into heat waves, severe thunderstorms, flash flooding, and winter storms, including interior snow. For each event, the observed no show proportion and AUROC were computed and expressed as a change relative to a baseline of all remaining days without severe weather events. Confidence intervals were computed following the same patient-level cluster bootstrap approach as before. As a negative control, we applied the same procedure to non-event days randomly sampled from months with fewer severe weather events (\ie, excluding June, July, and December).

\end{spacing}

\clearpage

% =========================
% Back matter
% =========================

\begin{spacing}{1.1}

\xhdr{Ethics approval} 
Parts of this work that relate to the use of the BEACON model at the University Hospital of Bern were approved by the Kantonale Ethikkommission, Canton of Bern (\#2025-00623).
Parts of this work that relate to tetanus progression prediction were approved by the Scientific and Ethics Committee of the Hospital for Tropical Diseases and the Oxford Tropical Research Ethics Committee (2261/QD-BVBND). Parts of this work that relate to sepsis quality reporting at UCSDH were exempted from institutional review board review following a determination that the work was quality improvement by the Aligning and Coordinating QUality Improvement, Research, and Evaluation (ACQUIRE) Committee (Project \#1403). Parts of this work that relate to the Clalit Health Services prediction model were approved by the Clalit Health Services Institutional Review Board (Helsinki) committee. Work conducted at Mount Sinai was approved by the Mount Sinai Institutional Review Board (STUDY-20-00338).

\xhdr{Data availability} The patient data and \name records analyzed in this study cannot be made publicly available because they contain protected health information. Requests for access to this data should be directed to Gunnar R\"{a}tsch (\href{mailto:raetsch@inf.ethz.ch}{raetsch@inf.ethz.ch}) for the Bern, Switzerland pilot; C. Louise Thwaites for the Ho Chi Minh City, Vietnam pilot (\href{mailto:louise.thwaites@ndm.ox.ac.uk}{louise.thwaites@ndm.ox.ac.uk}); Karandeep Singh for the San Diego, California pilot (\href{mailto:karandeep@health.ucsd.edu}{karandeep@health.ucsd.edu}); and Ankit Sakhuja and Benjamin Glicksberg for the New York, New York pilot (\href{mailto:ankit.sakhuja@mssm.edu}{ankit.sakhuja@mssm.edu}, \href{mailto:benjamin.glicksberg@mssm.edu}{benjamin.glicksberg@mssm.edu}). Severe weather event data was retrieved from the New York Significant Weather Events Archive from the U.S. National Weather Service, available at \url{https://www.weather.gov/okx/stormarchive}.

\xhdr{Code availability} The \name website is available at \url{https://medlogprotocol.ai}. Source code is available at \url{https://github.com/mims-harvard/MedLog}. 

\xhdr{Acknowledgements} 
We thank Eric Allman for valuable discussions regarding \texttt{syslog}, as well as David Blumenthal and Bakul Patel for critical review of this manuscript. We also thank Iñaki Arango and Lucas Vittor for assistance with the development of the API prototype. Finally, we thank the members of the 2024 Harvard Radcliffe Accelerator AIM Working Group on Bioethics and the Responsible AI for Social and Ethical Healthcare 2025 ``2.0'' Symposium for discussions that informed this work. We gratefully acknowledge the support of NSF CAREER 2339524, ARPA-H Biomedical Data Fabric (BDF) Toolbox Program, Harvard Data Science Initiative, Amazon Faculty Research, Google Research Scholar Program, AstraZeneca Research, Roche Alliance with Distinguished Scientists (ROADS) Program, Sanofi iDEA-iTECH Award, GlaxoSmithKline Award, Boehringer Ingelheim Award, Merck Award, Optum AI Research Collaboration Award, Pfizer Research, Gates Foundation (INV-079038), Chan Zuckerberg Initiative, John and Virginia Kaneb Fellowship at Harvard Medical School, Biswas Computational Biology Initiative in partnership with the Milken Institute, Harvard Medical School Dean's Innovation Fund for the Use of Artificial Intelligence, and the Kempner Institute for the Study of Natural and Artificial Intelligence at Harvard University. A.N. was supported by a Rhodes Scholarship, a Cosmos Fellowship, and a Y Combinator Summer Fellows Grant. I.S.K. was supported by OpenAI NextGenAI. D.A.C. was funded by an NIHR Research Professorship and an RAEng Research Chair, and supported by the Oxford NIHR Biomedical Research Centre. Figures~\ref{fig:elements}, \ref{fig:zurich}, \ref{fig:ucsdh}, and \ref{fig:overview} were created, in part, using BioRender (see \href{https://BioRender.com/6uscjoe}{https://biorender.com/6uscjoe}).

\xhdr{Competing interests} 
A.K. and V.N. are currently employed by Google DeepMind. D.D. is currently employed by e-Patient Dave, LLC. H.F.W. is currently employed by Healthcare Information and Management Systems Society, Inc. J.C.M. and P.L. are currently employed by Microsoft Research. J.R. is currently employed by E-Citizen Solutions Africa. S.H. is currently employed by Epic Systems Corporation. The other authors declare no competing interests.
\end{spacing}

\clearpage

% =========================
% References
% =========================
\section*{References}
\vspace{1em}
\begin{spacing}{1}
\printbibliography[heading=none]
\end{spacing}
\end{refsection} % end MAIN refsection

% =========================
% SUPPLEMENTARY INFORMATION
% =========================
\begin{refsection}

% --- SI divider page ---
\newpage
\begin{spacing}{1}
\setlength{\parskip}{12pt}
{\Large\bfseries\noindent\sloppy \textsf{%
  \begin{center}\Large{Supplementary Information for}\\[1mm]%
  \textbf{A global log for medical AI}\vspace{-10mm}\end{center}} \par}
{\noindent\sloppy \savedauthorblock}
\end{spacing}

\vspace{4em}
\begin{spacing}{1}
\noindent \textsbf{This PDF file includes:}
\begin{description}[labelsep=1em, leftmargin=2.2em, itemsep=1em]
  \item[] Supplementary Notes 1 to 4
  \item[] Supplementary Figures 1 to 2
  \item[] Supplementary Table 1
\end{description}
\end{spacing}

\clearpage

% --- SI content ---
\begin{spacing}{1}
\small
\renewcommand{\theHfigure}{SI.\arabic{figure}}
\renewcommand{\theHtable}{SI.\arabic{table}}
\renewcommand{\theHsection}{SI.\arabic{section}}
\captionsetup[figure]{labelformat=supplementary, labelsep=none}
\captionsetup[table]{labelformat=supplementary, labelsep=none}

%!TEX root = SI/SI-main.tex
\setcounter{section}{0}
\sloppy

% === Define macros for Supplementary content ===

% Supplementary Notes
\newcommand{\siNote}[2]{%
  \refstepcounter{section}% step section counter for referencing
  \vspace{2em}
  \section*{Supplementary Note \arabic{section}: #2}%
  \label{si:note:#1}%
}

% Supplementary Figures (use within figure environment)
% Usage: \siFigure{label}{Short caption}{Long caption with description}
\newcommand{\siFigure}[3]{%
  \caption[#2]{\textsbf{#2} #3}%
  \label{si:fig:#1}%
}

% Supplementary Tables (use within table environment)
% Usage: \siTable{label}{Short caption}{Long caption with description}
\newcommand{\siTable}[3]{%
  \caption[#2]{\textsbf{#2} #3}%
  \label{si:table:#1}%
}

\siNote{further-challenges}{Further operational considerations for \name}

\xhdr{Patient privacy and data security} Protecting patient privacy is paramount when implementing \name. Like EHR databases, \name systems will record identifiable protected health information and sensitive operational data. Access to \name records must therefore be tightly controlled within secure computing environments, and the same regulatory standards and security systems used to protect EHR databases -- Health Insurance Portability and Accountability Act (HIPAA), Health Information Technology for Economic and Clinical Health (HITECH) Act, General Data Protection Regulation (GDPR), or ISO/IEC 27001 \autocite{isotc_215_iso_2016, iso/iecjtc1/sc27ISOIEC270012022} compliance; role-based access control; audit logging; the use of pseudonymous identifiers; and storing content-addressed references rather than raw media -- should be adopted for storing \name records. For example, AI outputs that influence clinical decision-making can reside in the EHR, while pointers to these outputs can be included in \name records, excluding \name from the HIPAA-designated record set or legal medical record. 

Similar to how LLMs and other clinical foundation models are now fine-tuned and adapted within institutional firewalls, healthcare systems can deploy \name systems locally to prevent unauthorized access. Where privacy regulations permit, secondary analysis tools or federated algorithms can operate within these environments to de-identify \name records and compute aggregated performance metrics or summary statistics, allowing meta analyses while preserving patient confidentiality. Data sharing agreements could then support inter-institutional comparisons of anonymized \name records \autocite{shah_nationwide_2024}, especially to evaluate models deployed outside their training settings \autocite{yang_generalizability_2024}. For example, much as pharmacovigilance programs like the FDA Adverse Event Reporting System and the WHO Programme for International Drug Monitoring pool deidentified case reports to assess drug safety, deidentified or aggregated \name records could be shared with regulatory bodies tasked with post-market surveillance of medical AI systems \autocite{embi_algorithmovigilanceadvancing_2021}. However, these cross-institutional analyses will require privacy-preserving mechanisms for exchanging \name entries, such as secure multi‑party computation, homomorphic encryption, federated learning, or blockchain~\autocite{rieke_future_2020, xia_medshare_2017}.

\xhdr{Data storage and management} Capturing each AI interaction is necessary to ensure safety and accountability of medical AI, but it will generate substantial data volumes. Implementing \name will require investment in large-scale data storage, management, and networking infrastructure, similar to the investments made in developing EHR system. However, data storage and networking demands are not unique to \name systems. Healthcare systems around the world are projected to generate more than 10,800 exabytes of data annually by 2025 \autocite{banks_sizing_2020, telenti_treating_2020}, and a single hospitalization already produces approximately 150,000 discrete data elements \autocite{banks_sizing_2020, esteva_guide_2019}. The widespread adoption of AI only increases the urgency of building a robust and secure healthcare data infrastructure capable of transmitting and storing exabytes or zettabytes of data. To maximize the impact of \name, access to such infrastructure must be democratized so that all healthcare systems can participate in AI monitoring and improvement efforts.

In practice, organizations may default to retaining all \name events indefinitely to maximize observability and support retrospective analysis. However, as in other high-compliance domains, tailored strategies can balance safety, privacy, and cost. Institutions can adopt lifecycle-aware capture and retention policies that use full tracing during pilots and post-update periods, sampled or risk-triggered tracing in steady state, and tiered retention (\eg, long-term summaries with shorter-lived detailed artifacts). This approach preserves forensic and regulatory value while containing operational overhead.

\xhdr{Pathways to real-world deployment} \name can be adopted unilaterally within a health system to improve safety monitoring and evaluation with no external mandate. However, shared expectations and harmonized interfaces across EHR vendors, AI vendors, and health systems will reduce integration costs and enable consistent, multi-site analyses. We anticipate a mixed adoption pathway: early adopters implement \name locally, while emerging guidance from professional bodies and regulators fosters convergence toward uniform capture and exchange. To support legacy and \name-naive systems, organizations can implement \name at high-leverage points in the technology stack, \eg, as LLM proxies or API gateways that intercept AI calls, extract or augment metadata, and emit \name-compliant entries. Sidecar services can similarly wrap agent frameworks and tool calls to record inputs, retrieved context, outputs, and uncertainty estimates. These patterns accelerate adoption without waiting for deep vendor changes.

\name can be implemented using existing open standards and tooling. \name software should adopt the W3C PROV conceptual model for computational provenance; for example, the \name record and its fragments are \texttt{prov:Entity} instances; the model invocation itself is a \texttt{prov:Activity}; and the model and user are instances of \texttt{prov:Agent} or its subclasses, such as \texttt{prov:SoftwareAgent} or \texttt{prov:Person} \autocite{khalid_belhajjame_prov-o_2013}. For operational telemetry, \href{https://opentelemetry.io/}{OpenTelemetry} provides consistent schemas, collectors, and backends to transport and store event data across languages and platforms. For provenance tracking in, for example, the User identity field, \linelabel{line:si-operational-openlineage} \href{https://openlineage.io/}{OpenLineage} offers a framework for collecting lineage metadata. For clinical semantics and linkage, \href{https://hl7.org/fhir/}{Fast Healthcare Interoperability Resources (FHIR)} data formats and elements (\eg, AuditEvent, Patient, Condition, Observation, Practitioner, PractitionerRole) can anchor \name entries to standardized clinical entities \autocite{hl7_fhir_2023}. These interoperability layers enable scalable, vendor-agnostic deployments and lower the barrier to multi-institutional analyses.

\linelabel{line:si-audit-logs-start} \name also builds on EHR audit logs. EHRs often maintain event or audit logs for record access~\autocite{adler-milstein_ehr_2020, rule_using_2023, kannampallil_using_2023}. EHR audit logs demonstrate event-level logging in healthcare is feasible at scale: audit logs can capture time-stamped user actions, measure clinical work, and reveal aspects of care delivery that are not visible from clinical or claims data alone. For example, audit logs have been used to assess hospital workflow efficiency~\autocite{chen_inferring_2015}; identify interaction patterns associated with length of stay~\autocite{chen_interaction_2018} or timeliness of care~\autocite{noshad_context_2021}; detect clinical outliers~\autocite{bouarfa_workflow_2012}; study provider behavior~\autocite{amroze_use_2019}; and reveal contextual factors relevant to clinical process outcomes across multiple sites, even despite site-dependent documentation differences~\autocite{rose_team_2023}. These past studies of EHR auditing and logging can provide learnings for real-world deployment of \name beyond pilots reported in this study.\linelabel{line:si-audit-logs-end}

\xhdr{Global implementation of \name} For deployments in low- and lower-middle-income countries (LMICs), \name allows partial or incremental compliance. A minimal conformance profile can capture Header, Model instance, and Outputs, with other fields added as capacity grows. Local write-behind caching enables offline operation with delayed synchronization -- for example, between lightweight smartphone applications and a centralized \name server -- when connectivity becomes available. Where EHR systems or unique identifiers are limited, \name records can anchor to encounter-level metadata such as visit, time, location, and department, with optional FHIR linkage when feasible. \name can also map to widely used platforms such as OpenMRS \autocite{mamlin_cooking_2006} and DHIS2 \autocite{dehnavieh_district_2019}. 

In settings with limited infrastructure, lifecycle-aware retention and risk-triggered sampling will be essential. In countries with emerging regulatory frameworks, \name records can be overseen by health system administrators, health ministries, or international partners such as the World Health Organization. Funders should back pilot implementations across health systems to prevent widening disparities and to establish systematic monitoring of LLM outputs in patient care.

\xhdr{Aligning incentives and governance} The promise of \name depends on policy as much as technology. We must learn from past efforts to build distributed digital public health infrastructure \autocite{kadakia_modernizing_2021}. For example, although the HITECH Act of 2009 was successful in ubiquitizing EHRs, information exchange encountered roadblocks: in 2015, 96\% of hospitals either claimed exclusion from or did not report to specialized public health registries \autocite{office_of_the_national_coordinator_for_health_information_technology_hospital_2016}. To encourage the adoption of \name, the business case must be explicit: even without cross-institutional data sharing, \name delivers safety and quality improvement, liability management, and operational efficiency. Institutions can preserve competitive data advantages by retaining raw logs locally while participating in benchmarking consortia that return site-level insights as a reciprocal benefit. \name could even unlock new pathways to financial sustainability; for example, providers could leverage AI performance data to establish value-based contracts for AI services. Furthermore, to encourage clinician participation, deployments must build trust by presenting \name as a collaborative tool for enhancing clinical learning, addressing potential concerns about professional autonomy. To that end, establishing governance bodies with strong clinician representation is essential.

Incentives and risks for model developers and vendors must also be considered. For example, large numbers of input-output examples, uncertainty estimates, or reasoning traces from \name records could enable membership inference attacks that expose a model's training set \autocite{shokri_membership_2017, mattern_membership_2023} or extraction attacks to distill proprietary models into imitations \autocite{zhao_systematic_2025} or reconstruct proprietary prompting techniques or tool calls \autocite{sha_prompt_2024, zhang_extracting_2024}. To mitigate such adversarial misuse, which could discourage vendor participation, governance frameworks for \name should require technical safeguards and intellectual property protection agreements that prevent reverse engineering \autocite{he_cater_2022, li_llm-pbe_2024}. Certain \name fields, such as Model instance, Inputs, and Outputs, can also include data ownership tags to ensure clear provenance. Ultimately, success will hinge on mechanisms such as these to align incentives, governance structures, and funding models.

\siNote{further-discussion}{Further discussion on opportunities and impact of \name}

Large-scale logging of human-AI interactions has shown clear value for safety monitoring, jailbreak detection, usage analysis, benchmark construction, and instruction fine-tuning to align model outputs with human preferences \autocite{zhao_wildchat_2024, zheng_lmsys-chat-1m_2024}. The same applies in healthcare and medicine: monitoring AI through \name  can advance model development, evaluation, safety, reliability, and transparency. Below, we outline the opportunities for \name and the research, policy, and organizational changes needed to achieve them.

\xhdr{From human epidemiology to human-AI epidemiology}
Continuous monitoring of AI-human and AI-AI interactions establishes a data layer that has not previously existed in medicine. Systematic records of how AI systems participate in healthcare. Traditional epidemiology studies the distribution and determinants of health and disease in populations. Digital epidemiology expanded this scope by drawing on EHRs, claims, and patient-generated data. \name extends it further by treating AI itself as a measurable agent in clinical environments.  Epidemiology can begin studying machine behavior alongside human behavior. 

By capturing inputs, reasoning traces, outputs, and linked outcomes, \name records document how AI influences decision-making, clinical actions, and patient trajectories, transforming epidemiology from studying how humans and environments shape health to also studying how algorithms mediate those processes. Using implementation science approaches \autocite{longhurst_call_2024}, epidemiologists can analyze \name records to detect both positive \autocite{mcduff_towards_2025} and negative \autocite{budzyn_endoscopist_2025} performance changes with clinician-AI collaboration; variation in AI-human and AI recommendations across demographic groups, specialties, or regions~\cite{li2026scaling}; clinician over- or under-reliance on AI~\cite{tikhomirov2024medical}; workflow changes introduced by automation~\cite{ferber_towards_2026}; and the long-term effects of AI-assisted care on quality and safety~\cite{jacob2025ai}. 

These analyses will generate quantitative evidence for health policy, support population-tailored AI interventions, and guide precision medicine. \name systems give hospital quality-improvement teams operational intelligence across outpatient clinics, inpatient wards, operating rooms, and ancillary services, allowing analysts to link records to clinical and financial outcomes, identify patterns of risk, and refine AI workflows to improve safety and efficiency.

\xhdr{Real-time surveillance of medical AI safety}\linelabel{line:si-discussion-surveillance-start}
\name records give regulators the means to detect adverse events from near misses, errors, and model failures in real time, with event-level logs supplying the ``artifact collection'' that frameworks for medical algorithmic audits require \autocite{raji_closing_2020, liu_medical_2022}. Algorithmic auditing can be automated by drawing on patterns from modern cybersecurity monitoring, supervisor agents or other AI systems can process, summarize, and triage \name records, trigger alerts for additional human review, and make continuous organization-wide oversight operationally feasible \autocite{zhu_loghub_2023, qi_loggpt_2023, karlsen_benchmarking_2024}. 

Realizing this potential will require regulatory mandates for post-market reporting, such as those proposed by the U.S. Food and Drug Administration (FDA) \autocite{warraich_fda_2024} and the European Commission, alongside public-private partnerships such as a network of health AI assurance laboratories \autocite{shah_nationwide_2024}. Recent FDA guidance stresses credibility assessment plans, life cycle management, and automated oversight as conditions of regulatory approval \autocite{center_for_veterinary_medicine_considerations_2025, center_for_devices_and_radiological_health_artificial_2025}, requirements that cannot be met in practice without \name.\linelabel{line:si-discussion-surveillance-end}

\xhdr{Detecting dataset shifts in medical AI}
By logging AI-human interaction events, \name records allow developers to detect when model behavior deviates from expectations because of shifts in deployment data. Dataset shift arises when models encounter changes in patient demographics, clinical practices, medical technologies, or care settings compared to their training environments \autocite{subbaswamy_development_2020, finlayson_clinician_2021}. Event-level logging preserves the full context of each model invocation, including inputs, outputs, and outcomes, making it possible to track how shifts manifest across subgroups, workflows, or care settings rather than only at an aggregate level (Supplementary Note~\ref{si:clalit-casestudy}). 

The structure of \name records, which captures retrievals, generations, outcomes, and feedback, also allows us to distinguish between shifts in the underlying data distribution and shifts in how models are applied in practice. Comparing data distributions between training and deployment stages, using methods such as deep learning-based hypothesis testing \autocite{koch_distribution_2024, schrouff_diagnosing_2022}, enables early detection of global and subgroup-level shifts that standard outlier detection may miss \autocite{koch_hidden_2022}. \name records can also reveal performance degradation when AI models are retrained on new data, helping to identify risks such as data poisoning attacks \autocite{alber_medical_2025}.

\xhdr{Monitoring for bias in medical AI}
\name records systematically track model performance across attributes such as age, sex, race, ethnicity, socioeconomic status, and insurance coverage to assess bias \autocite{seyyed-kalantari_underdiagnosis_2021, daneshjou_disparities_2022, liu_translational_2023, xiong_how_2025}. Automated slice discovery methods applied to \name records surface differential performance across these groups \autocite{eyuboglu_domino_2022}, flagging cases where panels of clinicians and ethicists must determine whether observed differences reflect clinically justified variation or inequities that require remediation \autocite{liu_translational_2023, de_kanter_preventing_2023}. Sustained bias monitoring strengthens accountability, supports equitable AI use across diverse populations, and provide evidence to the regulators needed to evaluate whether AI models meet fairness and safety requirements.

\xhdr{Using \name data to improve AI models}
\name records capture real-world error cases, near misses, and uncertainty estimates that drive model refinement. Uncertainty estimates recorded for each prediction support active learning strategies, where the least confident predictions are flagged for expert review and used for model post-training \autocite{budd_survey_2021}. Beyond traditional supervised or reinforcement fine-tuning, \name traces power meta-learning strategies including curriculum learning \autocite{bengio_curriculum_2009, liu_competence-based_2023, rui_improving_2025}, lifelong learning \autocite{zheng_lifelong_2025}, and self-evolving agents \autocite{gao_survey_2025}. Drawing on feedback in \name traces, pre-trained models or agents can adaptively sequence prompt \autocite{liu_lets_2024} or data \autocite{bae_online_2025, chen_self-evolving_2025} examples, critique or revise their own actions \autocite{shinn_reflexion_2023, madaan_self-refine_2023}, generate new tasks from failures \autocite{qi_webrl_2025, zhou_self-challenging_2025}, update long-term memory modules \autocite{li_memos_2025}, or autonomously modify their tools or design \autocite{robeyns_self-improving_2025, hu_automated_2025}. 

De-identified \name records can also be mined to construct challenging, realistic benchmarks that outperform today's artificial evaluations at estimating real-world performance on clinical tasks \autocite{lin_wildbench_2024}. Machine learning teams embedded in healthcare systems can use \name records or \name-derived benchmarks to determine when to deploy, retire, or retrain AI models and compare the performance of different models (\eg, GPT-5 versus Claude Sonnet 4 versus Gemini 2.5 Pro) in real-time or retrospectively. Beyond model-level decisions, \name records can capture user interaction patterns, informing improvements in user interfaces, workflow integration, system management, and the development of new medical AI applications.

\xhdr{International evaluation of AI models} As \name records capture the same fields across sites, public health agencies and consortia can aggregate de-identified logs to benchmark AI models worldwide across geographic and economic strata. Medical AI research shows marked geoeconomic disparities: as of August 2024, only 2.3\% of studies were conducted in LMICs and only 6.3\% spanned more than one nation \autocite{yang_disparities_2024, han_randomised_2024}. Yet, early real-world LMIC deployments \autocite{mateen_trials_2025} have already reduced diagnostic and treatment errors \autocite{korom_ai-based_2025}. The opportunity to measure the global impact of clinical AI is therefore urgent and largely untapped. Routine, standardized logging can show whether AI tools narrow or widen performance gaps between resource-rich and resource-constrained hospitals, identify deployments that need additional context-specific training \autocite{yang_generalizability_2024}, and inform developers, regulators, and funders seeking to support systems that reduce rather than entrench global health inequities.

\xhdr{Advancing transparency for patients and clinicians}
\name records create traceable documentation of AI-generated content in EHRs and administrative claims. Clinicians can review the rationales behind AI outputs to guide patient care. In the United States, health data downloads are identified by the ``Blue Button'' logo \autocite{assistant_secretary_for_technology_policy_blue_2022, mandl_push_2020}; with wider \name adoption, exported health data could also include AI interaction traces. Patients using personal health LLMs \autocite{khasentino2025personal} may then rely on \name exports to audit the information available to their models or to transfer their digital medical assistants across platforms.

\siNote{limitations}{Limitations and challenges}

\linelabel{line:si-limitations-start} Below, we discuss limitations and challenges that can inform future implementations of \name.

\xhdr{Outcome linkage} \linelabel{line:si-limitations-linkage-start} Linkage of AI outputs to clinical actions and outcomes is often challenging~\autocite{joshi_ai_2025}. Many clinical outcomes occur days, months, or years after an AI interaction, and the relevant data may reside across disparate healthcare systems, registries, claims databases, or other sources. Even when an outcome is observable, attribution to a specific AI prediction may be ambiguous~\autocite{bleher_diffused_2022}. For example, a clinician may view an AI-generated recommendation but act for unrelated reasons; a patient may receive multiple AI-mediated messages across a care episode; or an AI alert and a clinician action may both be triggered by the same underlying deterioration rather than one causing the other. Thus, while \name records can establish temporal proximity, they may not be able to establish causal effect. Implementations of \name should therefore distinguish between different strengths of outcome linkage across multiple tiers of evidence. \linelabel{line:si-limitations-linkage-end}

\xhdr{Incomplete and biased capture} \linelabel{line:si-limitations-incomplete-start} If logging is incomplete, selectively enabled, or restricted to high-risk workflows, the resulting \name records may not represent the full breadth of AI use. While risk-triggered sampling and retention may reduce storage requirements, they may also bias data collection towards cases already recognized as unusual; meanwhile, low-risk interactions may be underrepresented, even though they could reveal slow-onset failures or long-term automation bias. In addition, AI use outside institutionally approved channels, such as consumer chatbots accessed on personal devices, would not be captured by \name. Therefore, analyses of \name records may underestimate the true scope of AI use. \linelabel{line:si-limitations-incomplete-end} 

\xhdr{Storage and network burden} \linelabel{line:si-limitations-storage-start} \name records may necessitate substantial storage, indexing, encryption, and backup (Supplementary Table~\ref{si:tab:syslog_medlog}). Generative AI and AI agents may exacerbate this footprint, as each user interaction can invoke many underlying model calls, tool invocations, or retrieval steps. To mitigate the data footprint and reduce the environmental impact, pointers instead of raw media can be stored; repeated artifacts can be compressed and de-duplicated; and storage can be tiered by risk or retention horizon. \name can also make compute costs more measurable by recording optional metadata such as token usage, compute cost, and storage size in the Header field. \linelabel{line:si-limitations-latency-start} Logging may also introduce latency; while this latency is likely to be minor, \name should be implemented such that \name record creation does not block model inference, alert delivery, or downstream clinical actions, for example, via asynchronous addition to incrementally construct a \name record (Figure~\ref{fig:elements}a). \linelabel{line:si-limitations-latency-end} \linelabel{line:si-limitations-storage-end} 

\xhdr{Agentic and multi-stage workflows} \linelabel{line:si-limitations-agentic-start} Agentic systems may decompose a task, query multiple tools, retrieve evidence, invoke other models, or update memory, making identifying a single model invocation challenging. \name addresses this challenge by linking records through shared identifiers, such as run IDs and parent event IDs. Advances in generative AI may also require updates or revisions to the \name protocol. Future work should demonstrate logging of generative or agentic AI systems with \name. \linelabel{line:si-limitations-agentic-end} 

\xhdr{Goodhart's law} \linelabel{line:si-limitations-goodhart-start} Once metrics are derived from \name records, developers or clinicians may optimize for logged indicators rather than improved patient outcomes. For example, systems could reduce recorded adverse events by narrowing logging criteria or changing documentation practices rather than changing human behavior or improving true model performance. Therefore, \name deployments should include pre-specified analysis and accountability plans. \linelabel{line:si-limitations-goodhart-end} \linelabel{line:si-limitations-end}

\siNote{casestudy}{AI monitoring for data drift detection}\label{si:clalit-casestudy}

To illustrate how \name can safeguard model performance, we present the following case study. 

Clalit Health Services deployed an AI model to predict hospitalization risk and prioritize chronic patients for proactive periodic nursing follow-up. Specifically, a gradient-boosting model with 48 features, including demographics, laboratory values, diagnoses, medications, and medical procedures, was trained to predict non-ambulatory hospitalization over one year of follow-up. Based on their predicted hospitalization risk, patients were prioritized for nursing follow-up every six months to two years. This model was embedded into Clalit's proactive-preventative interventions platform, C-Pi (\url{https://www.clalit-innovation.org/predictivemedicine}), which combines a decision-support system with an AI-based prioritization engine. C-Pi helps thousands of primary care physicians and community nurses identify patients who need proactive outreach and delivers detailed management recommendations to close care gaps across the population.

The hospitalization risk model was trained in 2020 using data from January 2018 to January 2019, and tested on a temporal hold-out set from January 2019 to January 2020. After demonstrating strong performance on benchmarks, it was deployed. Clalit implemented monitoring of the model's input features. In mid-2023, the monitoring system detected a distribution shift in the ``Lactate Dehydrogenase Last Value (LDH)'' feature. A subsequent investigation revealed that this distribution shift was caused by a centralized switch to a new LDH testing kit in March 2023. After March 2023, LDH testing was conducted with the new kit, gradually altering the distribution of LDH values. This real-world instance of data drift was detected by the AI monitoring system at Clalit.

Supplementary Figure \ref{si:fig:casestudy} depicts the shift in LDH distributions across time. Starting from the training period (January 2018), remaining stable until the kit change (March 2023), and then gradually diverging over the next 18 months. We  simulated the impact of the drift on hospitalization risk predictions by running the model retrospectively at 3-monthly intervals from June 2023 through September 2024 both with and without correction for the shifted LDH values. 

Even small feature drifts could propagate into clinically meaningful prediction errors: by 18 months, nearly 10\% of patients would have had the absolute risk scores shifted by $>$ 0.1\%, and about 1\% by $>$ 1\%. If the change had occurred in a feature with increased feature importance, this data drift could have affected the final predictive score. This case study demonstrates how continuous monitoring of AI can detect subtle, system-level changes that would otherwise degrade model performance.
\clearpage

%!TEX root = SI/SI-main.tex
% \newgeometry{top=1cm, bottom=2cm, left=2.5cm, right=2.5cm}
\setcounter{figure}{0}
\clearpage

\begin{figure}[!t]
  \centering
  \includegraphics[width=\textwidth]{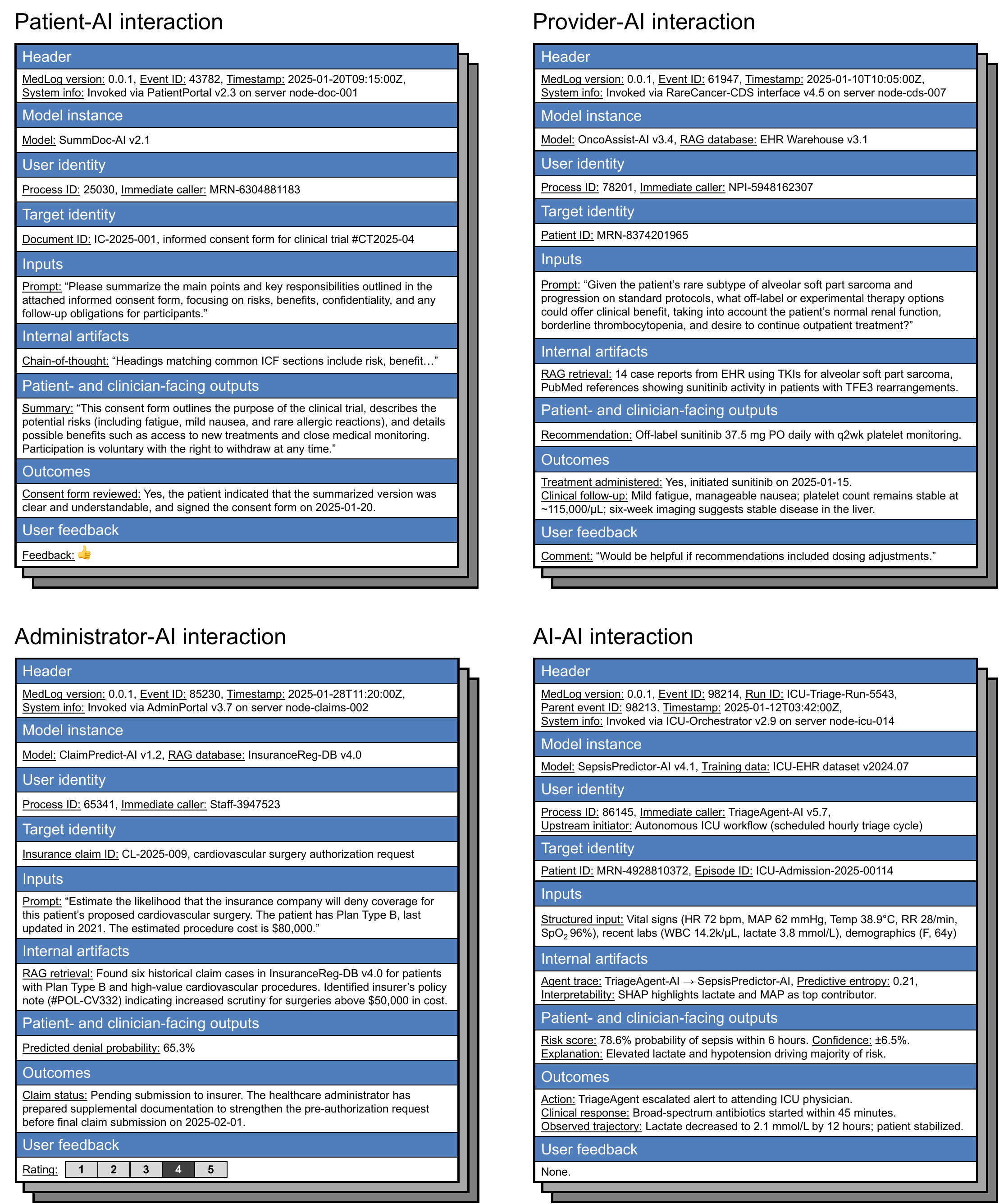}
  \caption{\textsbf{Examples of \name records.} Examples correspond to scenarios in Figure \ref{fig:elements}.}
  \label{si:fig:examples}
\end{figure}
\clearpage

\begin{figure}[!t]
  \centering
  \includegraphics[width=\textwidth]{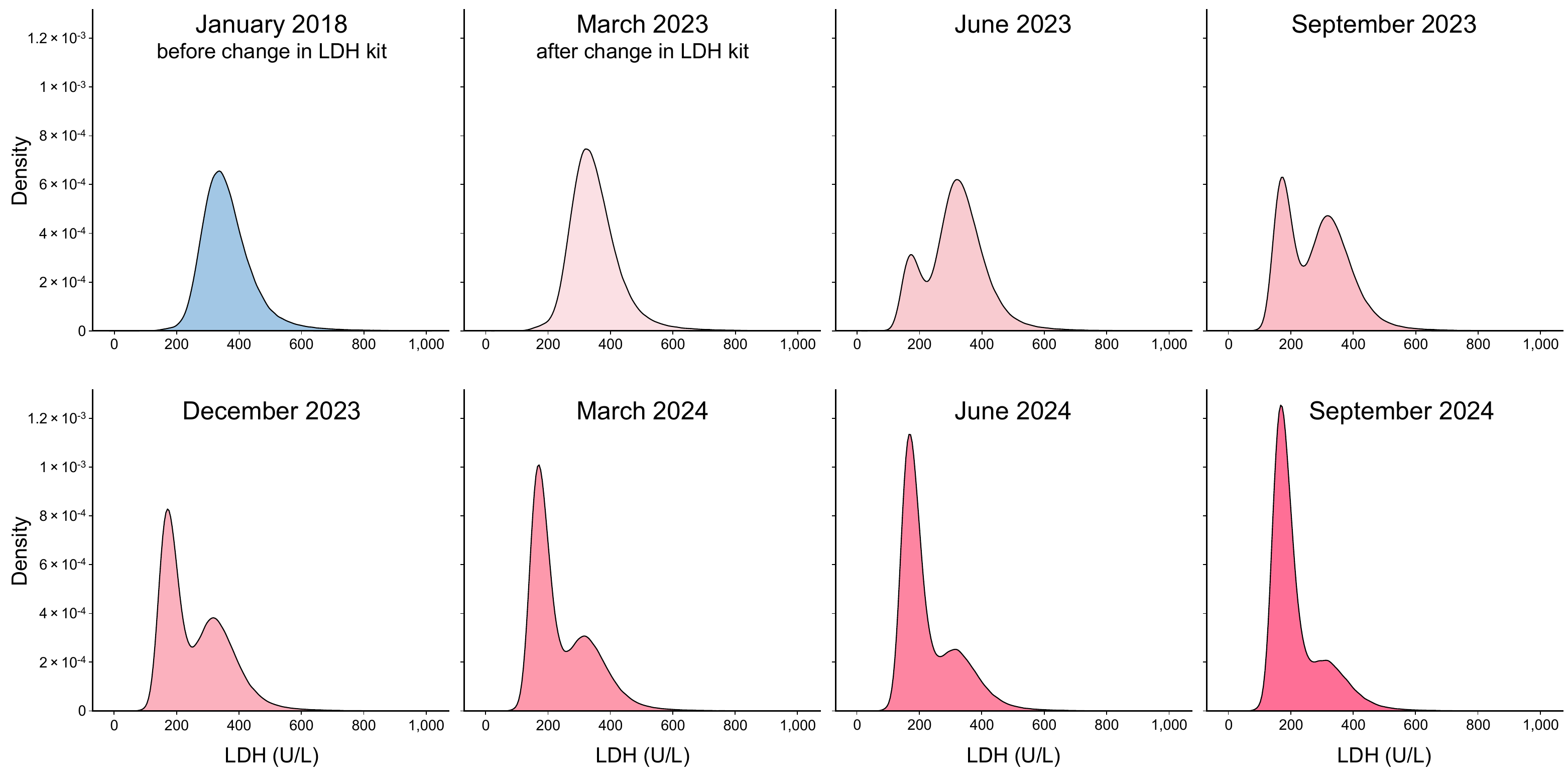}
  \caption{\textsbf{AI monitoring detects real-life data drift.} Density plots show the distribution of the ``Lactate Dehydrogenase Last Value (LDH)'' feature during the training period (January 2018), immediately after the test kit change (March 2023), and in subsequent quarterly snapshots through September 2024. The introduction of the new test kit caused a gradual shift in the distribution of LDH values, which was automatically detected by the AI monitoring system.}
  \label{si:fig:casestudy}
\end{figure}
\clearpage
\clearpage

%!TEX root = SI/SI-main.tex
% \newgeometry{top=1cm, bottom=2cm, left=2.5cm, right=2.5cm}
\setcounter{table}{0}
\clearpage

% Brand colours (tweak to taste)
\definecolor{HeaderBG}{HTML}{003366}   % deep navy
\definecolor{HeaderFG}{HTML}{FFFFFF}   % white
\definecolor{AttrBG}  {HTML}{E3E8F0}   % very light slate
\definecolor{RowAlt}  {HTML}{F7F9FC}   % ultra-light gray
\arrayrulecolor{gray!55}                % light gray rules

% Shorthand for left-column cells
\newcommand{\attrcell}[1]{\cellcolor{AttrBG}\textbf{#1}}
\newcommand{\tightrule}{\specialrule{\lightrulewidth}{0pt}{0pt}}
\setlength{\tabcolsep}{6pt} 

% Define a centered X column type
\newcommand{\head}[1]{\multicolumn{1}{c}{\textbf{#1}}}

% SUPPLEMENTARY TABLE 1
\begin{table}[ht]
  \centering
  \rowcolors{3}{RowAlt}{white}  % alternate rows starting with the 3rd
  \begin{tabularx}{\linewidth}
  {@{\hspace*{6pt}}                        % pull left edge back 3 pt
    >{\raggedright\arraybackslash}p{3cm}
    X X
   @{\hspace*{6pt}}} 
    % --- header ---
    \rowcolor{HeaderBG}
    \textcolor{HeaderFG}{\textbf{Attribute}} &
    \textcolor{HeaderFG}{\textbf{\texttt{syslog}}} &
    \textcolor{HeaderFG}{\textbf{\name}} \\[0.2em]
    \tightrule
    % --- data rows ---
    \attrcell{Purpose} &
      A protocol to send event messages from network devices and applications to a centralized logging server &
      A protocol for event-level logging of clinical AI
      \\ \tightrule
    \attrcell{Users} &
      IT and security teams &
      Clinicians, AI/ML engineers and researchers, safety regulators \\ \tightrule
    \attrcell{Structure} &
        \vspace{-1em}
        \begin{enumerate}[left=0pt]
          \item \texttt{HEADER}, which includes \texttt{PRI} (facility and severity), \texttt{VERSION} (version of the \texttt{syslog} protocol), \texttt{TIMESTAMP}, \texttt{HOSTNAME} (hostname and the domain name of the originator), \texttt{APP-NAME} (device or application that originated the message), \texttt{PROCID} (used to detect discontinuities or group messages), and \texttt{MSGID} (identifies the type of message)
          \item \texttt{STRUCTURED-DATA}, which can contain multiple structured data elements (\texttt{SD-ELEMENT}), each recorded as a name (\texttt{SD-ID}) and a parameter name-value pair (\texttt{SD-PARAM})
          \item \texttt{MSG}, a free-text message
        \end{enumerate} &
        \vspace{-1em}
        \begin{enumerate}[left=0pt]
          \item Header
          \item Model instance
          \item User identity
          \item Target identity
          \item Inputs
          \item Internal artifacts
          \item Patient- and clinician-facing outputs
          \item Outcome
          \item User feedback
        \end{enumerate} \\ \tightrule
    \attrcell{Granularity} &
      One record per event &
      One record per model invocation \\ \tightrule
    \attrcell{Privacy} &
      Clear-text protocol with no default encryption; not designed for sensitive data &
      Contains protected health information; encryption required \\ \tightrule
    \attrcell{Storage} &
      Moderate (\(\gtrsim\) KB-GB/day per hospital) &
      Large (\(\gtrsim\) GB-TB/day per hospital) \\
    \tightrule
  \end{tabularx}
  \caption{Design and feature comparison of \texttt{syslog} (RFC 5424 \autocite{gerhards_syslog_2009}; for the legacy BSD \texttt{syslog} protocol specification, see RFC 3164 \autocite{lonvick_bsd_2001}) and \name.}
  \label{si:tab:syslog_medlog}
\end{table}
\clearpage
\end{spacing}

\clearpage

% --- SI references  ---
\section*{References}
\vspace{1em}
\begin{spacing}{0.95}
\printbibliography[heading=none]
\end{spacing}
\end{refsection} % end SI refsection

\end{document}